\documentclass{article}

\usepackage{arxiv}

\usepackage[utf8]{inputenc} 
\usepackage[T1]{fontenc}    
\usepackage{hyperref}       
\usepackage{cite}
\usepackage{url}            
\usepackage{booktabs}       
\usepackage{amsmath,amssymb,amsfonts}       
\usepackage{nicefrac}       
\usepackage{microtype}      
\usepackage{lipsum}

\usepackage{algorithm}
\usepackage{algpseudocode}
\usepackage{multirow}
\usepackage{listings}

\usepackage{caption} 
\usepackage{graphicx}
\usepackage{textcomp}
\usepackage{xcolor}
\usepackage{pict2e}
\graphicspath{ {./figures/} }

\title{
\vspace{-3.5em}\footnotesize This work has been submitted to the IEEE for possible publication.\\
Copyright may be transferred without notice, after which this version will no longer be accessible.\\\vspace{4.25em}
\LARGE \bf
Approach to Finding a Robust Deep Learning Model
}

\author{
 Alexey Boldyrev \\
  Laboratory of Methods for Big Data Analysis\\ HSE University, 101000 Moscow, Russia\\
   \And
 Fedor Ratnikov \\
 Laboratory of Methods for Big Data Analysis\\ HSE University, 101000 Moscow, Russia\\ and\\ Yandex School of Data Analysis\\
  \And
 Andrey Shevelev \\
  Laboratory of Methods for Big Data Analysis\\ HSE University, 101000 Moscow, Russia\\
}

\begin{document}
\maketitle
\begin{abstract}
The rapid development of machine learning (ML) and artificial intelligence (AI) applications requires the training of large numbers of models. This growing demand highlights the importance of training models without human supervision, while ensuring that their predictions are reliable. In response to this need, we propose a novel approach for determining model robustness. This approach, supplemented with a proposed model selection algorithm designed as a meta-algorithm, is versatile and applicable to any machine learning model, provided that it is appropriate for the task at hand. This study demonstrates the application of our approach to evaluate the robustness of deep learning models. To this end, we study small models composed of a few convolutional and fully connected layers, using common optimizers due to their ease of interpretation and computational efficiency. Within this framework, we address the influence of training sample size, model weight initialization, and inductive bias on the robustness of deep learning models.
\end{abstract}

\keywords{machine learning \and deep learning \and convolutional neural networks \and model selection \and model robustness \and inductive bias \and generalization to unseen data \and automated learning}

\section{Introduction}
\label{sec:introduction}
The impressive successes of using deep learning in machine learning (ML) tasks are due in part to the existence of estimates for a sufficient architecture of a fully connected (FC) network with Sigmoid~\cite{cybenko1989approximation} and ReLU activation functions~\cite{hanin2017approximating}, for which solutions for continuous target functions have been obtained.
However, in the case of target functions for which their continuity is unknown, such an estimate does not exist at the time of writing.

The training of a neural network model depends on a number of conditions.
The behavior of gradient descent strongly depends on the chosen learning rate.
Too small values of the learning rate lead to long learning time, too large values lead to the instability of learning.
The training behavior of the neural network also depends on other hyperparameters used, such as the batch size.

When training deep neural networks, researchers have to select their hyperparameters. Due to the number of hyperparameters to be selected and the duration of training a neural network, the cost of training neural networks with reliable predictions is multiplied.
In addition to the standard hyperparameters like learning rate and batch size, the parameters of hidden layers are optimized as well, for example, the number of neurons and the size of filters in convolutional layers in case of Convolutional Neural Network (CNN) architecture.
The paper~\cite{hinz2018speeding} addresses the challenge of optimizing hyperparameters for deep CNNs, which is typically a computationally intensive process. The authors compare various state-of-the-art hyperparameter optimization algorithms and propose efficient strategies and methods to accelerate this optimization by initially employing lower-dimensional data representations and then progressively increasing the input dimensionality as the optimization progresses.

Neural Architecture Search (NAS), which is one of the promising topics in machine learning, allows researchers to automate the entire design of neural networks to achieve better performance on the given task and training sample~\cite{ren2021comprehensive, chitty2023neural}. Using benchmarks in NAS enables performing architecture searches that are cost-effective to evaluate.
Such benchmarks exist, for example, for a number of tasks in computer vision~\cite{ying2019bench, dong2020bench} and for natural language processing~\cite{klyuchnikov2022bench}.

There are tasks for which it is necessary to obtain solutions that are close to each other in terms of loss, regardless of the input data (taken from one population) and the initialization of weights.
This kind of stable learning is required, for example, for automated machine learning (AutoML).
The combination of neural architecture search and hyperparameter optimization is a hot topic in AutoML~\cite{he2021automl}.
Automated learning is used in optimization loops where the human verification of the neural network prediction is replaced by a procedure of selecting a solution and passing it to the next optimization steps.
For example, the design of complex scientific equipment can use such optimization loops with automated ML~\cite{dorigo2023toward}.

A number of papers describe finding robust neural networks when solving certain problems.
It is worth noting that the term ``robustness'' denotes different phenomena in different works.
Following Huber's definition~\cite{huber2011robust}, robustness signifies insensitivity to small deviations from the assumptions.
In this context, we are not only interested in how outlier resistant the model under consideration is, but also in how exactly the model is trained on different sized training samples.
It is assumed that there is a high quality random split between the training and test samples, i.e. outliers are similarly distributed in the subsamples.
In our study, a robust model is one that produces close losses regardless of the choice of a fixed-size training sample from the population.
The nature and size of the input data for training can influence the difficulty of finding a solution, especially a robust one.
There is a minimum required amount of data in the training sample to obtain robust predictions on the test sample~\cite{foody2006training}.
In computer vision tasks, the scale of objects is another factor affecting the robustness of predictions in fully connected and convolutional neural networks.
In the paper~\cite{barisin2024riesz}, the authors presented a novel scale-invariant neural network built on the Riesz transform. This network inherently generalizes to previously unseen data and adapts to objects of arbitrary scales.
Another method of handling objects of different scales is provided by Vision Transformers (ViTs)~\cite{dosovitskiy2020image}. Each input image is divided into patches, and regardless of the patch size, the transformer processes each patch along with its positional information, allowing it to identify and relate features across the entire image.

The paper~\cite{he2015delving} introduces a new weight initialization scheme, known as He (or Kaiming) initialization, specifically designed for networks using ReLU activations, which helps address the issues of exploding and vanishing gradients.
By applying these techniques and enhancements to deep CNN architectures, the researchers achieve significant improvements in performance, surpassing human-level accuracy on the ImageNet classification benchmark~\cite{deng2009imagenet}.
The problem of robust training and the influence of weight initialization is addressed in the paper~\cite{cyr2020robust}.
The authors analyzed a potential failure mode for some initializations that leads to an initial basis with strong linear dependence, and demonstrated this failure for Glorot and He initializations, which are commonly used in conjunction with ReLU activation functions. The authors proposed the ``box initialization'' method for deep ReLU networks that achieves relatively stable convergence in some problems.
The effect of weight initialization on the stability of the model solution is shown in paper~\cite{he2016deep}.
The paper~\cite{zhu2022robustness} is closely related to our study and discusses the effect of initialization of model weights on its robustness on theoretical grounds.
The authors introduce a definition of perturbation stability that can describe the robustness of a deep learning model, and conclude that perturbation stability under He initialization increases with depth (i.e., average robustness decreases).
The authors consider a two-layer ReLU network in the underparameterized setting and conclude that neural network width always helps to increase model robustness.
The authors also demonstrate the underlying rationale for the robustness of overparameterized neural networks, which may prove advantageous.
The authors of paper~\cite{schaeffer2023double} list factors for which double descent is not observed after switching between underparameterized and overparameterized regimes.

In the paper~\cite{colbrook2022difficulty}, the authors discuss the balance between stability and accuracy in existing deep learning methods for image reconstruction. The authors demonstrate the main well-conditioned problems in computing where deep neural networks with good approximation properties can be proven to exist; but argue that there is no algorithm, even randomized, that can train such a neural network.

The authors of the paper~\cite{haber2017stable} discuss the learning stability of ResNet networks, justifying why the widely used forward propagation scheme in deep learning may be unstable. They analyze these networks using concepts from numerical analysis and dynamical systems. The authors propose to view deep neural networks as discrete dynamical systems, where stability can be improved by drawing on methods used to solve ordinary differential equations. This perspective helps explain why ResNets exhibit more stable forward propagation, which explains their superior performance in training deep models.

The authors of the paper~\cite{hendrycks2021many} discuss the controversial notions of how best to assess robustness, and propose several additional
several additional data sets to address uncovered aspects of distributional shifts and model performance degradation that have not been well studied.

By reasoning about inductive biases, the authors of the paper~\cite{goyal2022inductive} show the connections between state-of-the-art deep learning and human-level cognition. The authors state that given a training sample of finite size, inductive bias is the only way to generalize to new input configurations. Inductive bias may not only be expressed in terms of a model architecture, but may also be related to the way the network is trained.

In the paper~\cite{koh2021wilds} a benchmark for developing and evaluating algorithms for training models that are robust to distribution shifts is proposed.
The authors conclude that developing models capable of maintaining robustness against distribution shifts in real-world scenarios continues to be an unresolved challenge.

The authors of paper~\cite{schneider2020improving} discuss the vulnerability of state-of-the-art computer vision models to image corruptions like blurring or compression artefacts.
The authors infer that the observed performance decline on corrupted images can largely be attributed to the variation in feature-wise first and second order moments.

The paper~\cite{bai2021recent} discusses adversarial robustness. The authors conclude that robustness in adversarially trained models is not guaranteed due to inadequate methods for solving non-convex min-max problems and due to severe overfitting, which is common in adversarial training.
The authors of the survey on Adversarial Robustness~\cite{silva2020opportunities} distinguish three categories of robust optimization: adversarial (re)training, regularization approaches, and certified defenses. In addition, the authors conducted a survey of methods that formally establish robustness certificates, either by solving the optimization problem exactly or by approximations using upper or lower bounds.

Despite the numerous implementations of robust neural networks for specific applications, a comprehensive study of robustness in deep learning is still lacking.
In this paper, we propose a novel approach to determine model robustness. This approach consists of the following steps: a certain number of model instances are selected, where all instances have identical architecture and hyperparameters.
These instances are then trained on different training samples and with different random initialization of the model weights, if applicable to the model.
The measure of robustness is determined by an appropriate statistical measure of the set of test losses of model instances, when a specified number of iterations is reached or when an early stopping criterion is met.

It is important to note that a robust model identified by this approach does not necessarily have to have good performance.
For example, a constant model will be robust because all its instances predict the same constant regardless of the training sample.
If the test sample is sufficiently large, the test loss of all constant model instances will be roughly similar.
To address this limitation, we enhance the proposed approach for determining model robustness with a novel model selection algorithm.
Thus, we can select the robust models among the top-performing ones.
This approach can be regarded as an enhancement to existing AutoML methods, offering a new automated model training strategy.
A potential application of this approach is the automated training of models for fault-tolerance systems, since our approach allows arbitrary evaluation of the closeness of the performance of a given number of instances of each model from the presented set.
In this respect, our empirical robustness approach extends the approaches discussed in the survey~\cite{silva2020opportunities} to certify the robustness of deep learning models by using an appropriate estimate of the model's loss variability given the required number of instances of the model.

In this study, we apply the robustness determination approach to deep learning models, focusing on those with a limited number of convolutional and fully connected layers. This choice stems from several factors. Primarily, our objective is to identify a stable deep learning model for a specific problem that offers a comparable performance to the baseline. Secondly, to study the stability of large sets of instances of the chosen model, we need to train from scratch several tens to thousands of instances of the model, depending on the experiment. Therefore, we prefer fast learning models with a small number of layers and not a very large number of parameters.

In the present study, we employed the robustness determination approach to deep learning models in multiple ways.
First, we narrowed down the class of models we would consider next.
To do so, we sequentially selected the activation functions and optimizers that provided the most stable performance for models with two convolutional and two fully connected layers on a small set of hyperparameters.
We then extended the set of architectures by specifying a possible range of the number of fully connected layers and varying the hyperparameters of the layers.
To the resulting set of models (a total of 6\,912 models for each of the two tasks), we applied the model selection algorithm mentioned above, which selects the most robust models among the best-performing models.
We repeated the robustness determination approach for the two models selected by the model selection algorithm in each of the tasks.
In these experiments, we use a larger number of instances of each model to investigate in detail the effects of training sample size and initialization of weights on model robustness.
Training different instances of the same model on samples of different sizes allows us to see how the model is trained in under- and over-parameterized regimes and thus verify the conclusions of the authors of the article~\cite{schaeffer2023double}.
For the selected models, we also investigate how robust they are to changes in the distribution by first selecting models on a simpler dataset and then observing their performance on a more complex one.

For a more comprehensive comparison and to validate the proposed robust model selection method, we compared the performance of the selected robust model with the best models identified using the Bayesian optimization-based NAS approach~\cite{watanabe2023tree}. The implementation details of the NAS approach and the results of this comparison are presented in Appendix~\ref{NAS_comparison}.

In our research we assess the implications of introducing inductive bias into the model by comparing its behavior with and without additional information on datasets of different complexity.
Following the suggestion of the authors of the paper ~\cite{goyal2022inductive} to investigate the influence of high-level variables on the learning of latent attribute representations, we have identified two inductive biases that play a causal role and can be easily introduced in the dataset we used due to its nature.

Finally, both the model robustness approach and the model selection algorithm proposed above are meta-algorithms and can be used with a set of arbitrary machine learning models that need only be relevant to the task.

\section{Dataset}
\label{sec-1}
\begin{figure*}[h]
    \centering
    \includegraphics[height=0.27\textwidth]{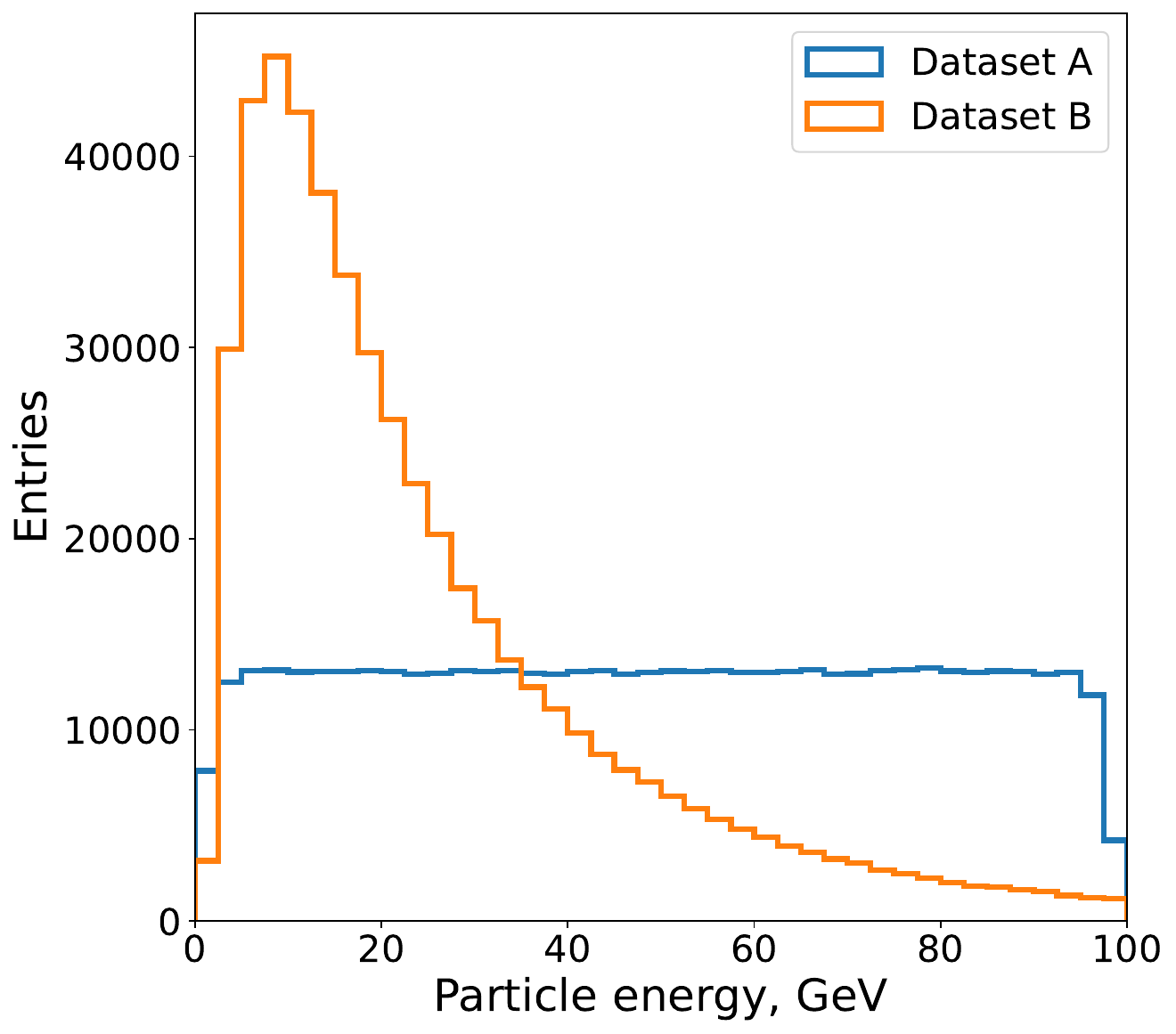}
    \hspace*{0.3em}
    \includegraphics[height=0.27\textwidth]{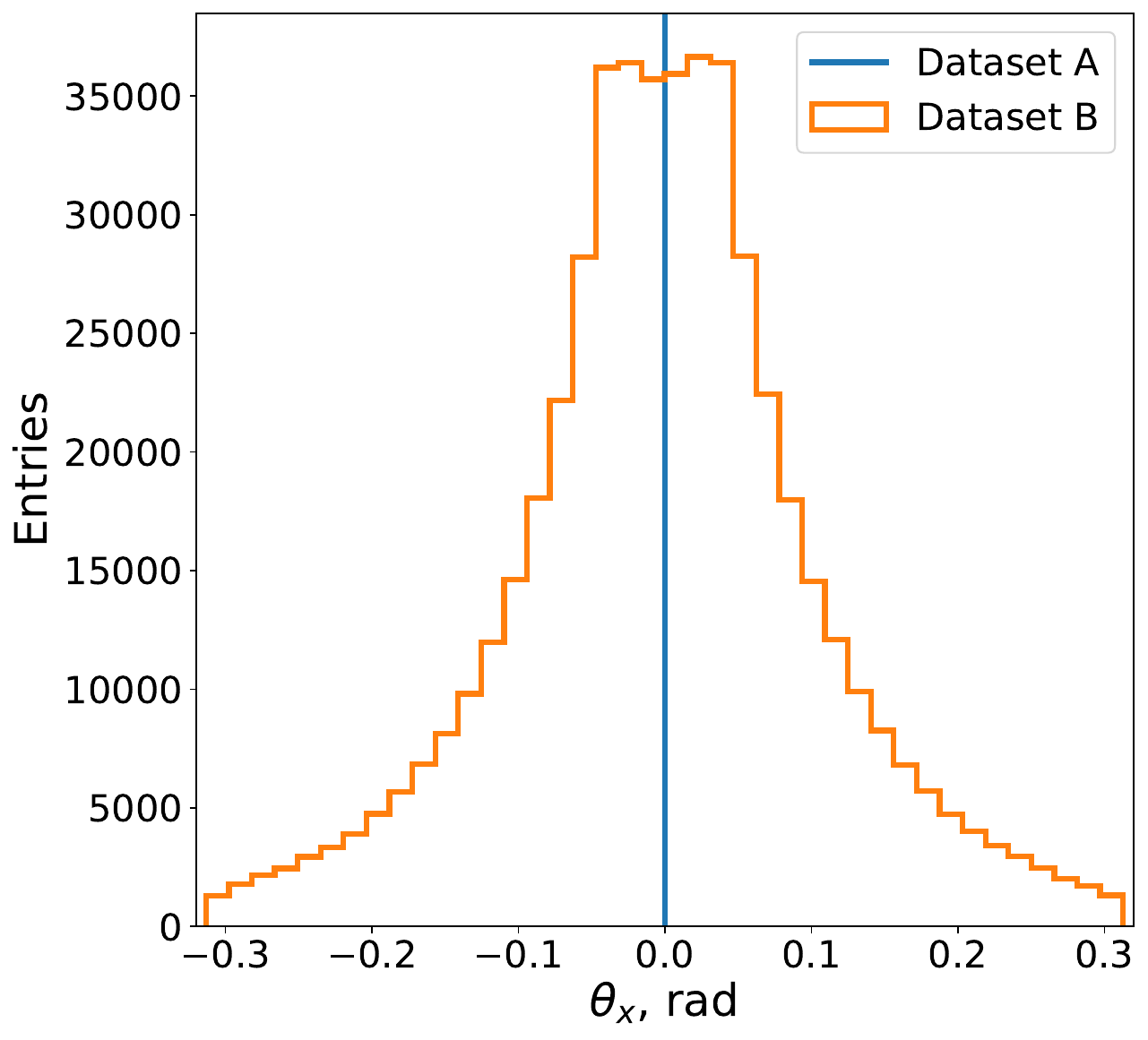}
    \hspace*{1.6em}
    \includegraphics[height=0.275\textwidth]{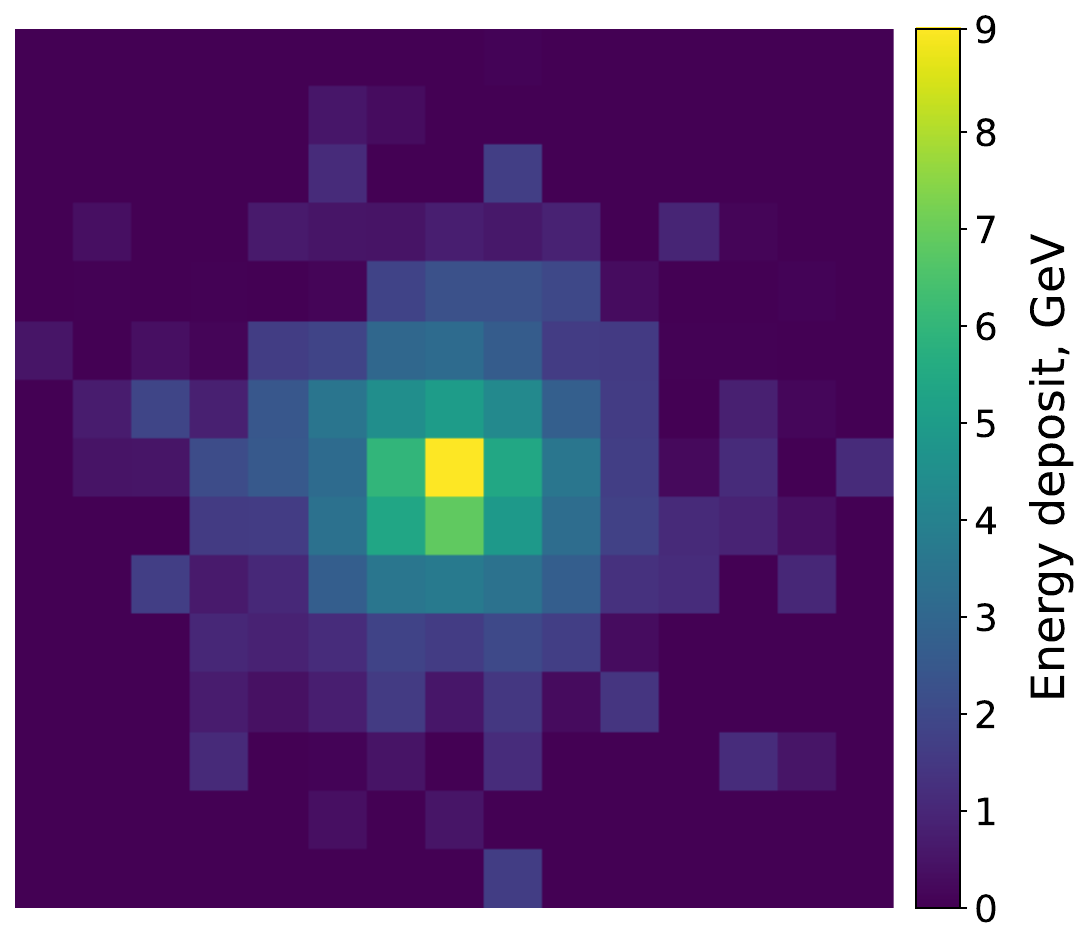}
    \caption{Properties of input data for Dataset A and Dataset B. Left: A histogram of energy spectrum. Center: A histogram of angular spectrum on one of the axes of the calorimeter plane. Right: Example of a calorimetric cluster from a particle with energy 65 GeV.}
    \label{fig_dataset_properties}
\end{figure*}

The input data consist of the properties of a particle entering the calorimeter and its response to it.
In this study, the particles are single photons. We set the energy, the position, and the incident angle of the particle approaching the calorimeter.
For every particle independently, these properties are passed to the \textsc{GEANT4} simulation toolkit~\cite{agostinelli2003geant4}, which produces a response from such a particle in its encoded model of a particular calorimeter.
The simulation model reproduces laterally segmented Shashlik-type electromagnetic calorimeter~\cite{barsuk2000design} similar to that used in the LHCb experiment~\cite{alves2008lhcb} at CERN.
The response to a particle in such a model is described by a $15 \times 15$ matrix, each element of which corresponds to the energy deposit in the corresponding cell of the calorimeter.
The set of energy deposits in such a matrix is hereinafter referred to as a calorimetric cluster, an example of which is shown in the right panel of Fig.~\ref{fig_dataset_properties}.
For the chosen calorimeter technology, the size of the cluster is proportional to the energy of the particle that generated it.

Two datasets are used: a simplified one, in which all particles approach the calorimeter normal to its surface and their energy is distributed uniformly from 1 to 100 GeV (referred as ``Dataset A''), and a realistic one, in which the incident angle of particles in the calorimeter varies in the range $|\theta_{x,y}| < 0.3$ (in radians), and their energy is related in some way to the angle of incidence, and it is distributed almost inversely exponentially in the range from 1 to 100~GeV (referred as ``Dataset B'').
For the reasons listed above, Dataset B is more complex and we used it as a test to generalize the models that performed best on Dataset A.
We created these two datasets using \textsc{GEANT4} simulation toolkit\footnote{The details of the simulation setup are described in~\cite{boldyrev2020ml}}.
Examples of particle energy and incident angle distributions used in the simulation are shown in the left and the center panels of Fig.~\ref{fig_dataset_properties}, correspondingly.
The arrival position of the particles in the calorimeter is uniformly smeared over the central cell for both datasets.
The calorimeter response can be represented as an image of the calorimeter cluster.
Each observation contains 225 features corresponding to the energy deposited in each of the $15 \times 15$ cells of the calorimeter.
An example of an observation is shown in the right panel of Fig.~\ref{fig_dataset_properties}.
Each feature (a pixel in calorimetric cluster) takes values of real numbers in a range from 0 to 14~(50)~GeV for the Dataset A (Dataset B), as shown in the corresponding distributions in Fig.~\ref{fig_feature_values}.
The data are not processed in any way, each model takes only raw feature values (energy values in the calorimeter cells) as input, except for models that also use the sum of the energies in all cells (one additional feature) and models that use the position of the center of mass of all cells (two additional features).
Both datasets contain 500\,000 observations each and are stored in the ROOT files~\cite{brun1997root}.
For both datasets, we randomly split the entire sample in a ratio of 0.5/0.5, and then we sample subsamples from one half for model training and model selection, and use the other half to test model performance.
We load the datasets using the Uproot~\cite{pivarski2017uproot} library and pass it as a tensor to PyTorch~\cite{paszke2019pytorch}.

\section{Method}
\label{sec-2}

\subsection{Baseline Approaches}\label{sec-3.1}
The main results of this paper are compared with the predictions of the following baseline approaches:
\begin{itemize}
    \item Scaled sum of energy deposits in all calorimeter cells to determine the particle energy;
    \item Barycenter of energy deposits in all cells convolved with an S-shaped curve to obtain the particle position\footnote{Examples of the use of S-shaped curve and its correction are discussed in~\cite{awes1992simple, schwindling1998s, sacco2003position}}.
\end{itemize}

These rather simple fully parametric models are widely used in calorimeter R\&D in high-energy physics, and are well suited for evaluating the quality of a single particle reconstruction in such detectors~\cite{fabjan2003calorimetry}. The source of stochasticity in the predictions of such models is the finite energy resolution of the calorimeter. In addition to these baseline models, some of the results discussed below are compared with the predictions of regularizing gradient boosting models applied to this problem previously by our group~\cite{boldyrev2020ml}. To ensure that the models are correctly compared, the parametric approaches and the gradient boosting models are trained on the same training sample as used by the neural network approaches discussed below. For a proper comparison, the predictions of all models considered in this paper are done on the same test sample or its subsamples.

\subsection{Loss Functions}\label{LossFunctions}
Particle reconstruction consists of estimating particle parameters such as energy and position in the plane of the electromagnetic calorimeter.
Thus, possible loss functions could be  $\mathrm{RMSE}(E)$ for energy estimation, and $\mathrm{RMSE}(x)$, $\mathrm{RMSE}(y)$ and $\mathrm{RMSE}(\sqrt{x^2 + y^2})$ for position estimation, respectively.
In both Dataset A and Dataset B, the dependence of the variability of the estimated energy on its value is observed.
Thus, using $\mathrm{RMSE}(E)$ as a quality metric for energy, the same error in energy determination will be negligible at high energies and significant at low energies.
To mitigate this effect, we use the following normalization in the quality metric evaluation for the energy:
$$\sqrt{\overline{\Bigg(\frac{E_{pred} - E_{true}}{E_{true}}\Bigg)^2}},$$
where $E_{pred} (E_{true})$ is predicted (true) energy, respectively.
This allows us to make the quality of energy reconstruction of low-energy particles and high-energy particles comparable and in this case the model is not specialized to determine the energy of particles of the most represented energy range.
Using the loss function $\mathrm{RMSE}(E)/E$ instead of $\mathrm{RMSE}(E)$ partially equalizes the stochasticity of the target variable.
Therefore, all experiments in this paper that address the energy reconstruction problem utilize the $\mathrm{RMSE}(E)/E$ loss function, unless otherwise stated.

To evaluate the quality of the position reconstruction, the $\mathrm{RMSE}$ of one of the coordinates is used, $x$, since the data has radial symmetry and the problem of determining the other coordinate is equivalent. This simplifies the problem a bit and has a clearer and more efficient physical meaning than considering $\mathrm{RMSE}(\sqrt{x^2 + y^2})$. 

\subsection{Preliminary Experiments}
Since the calorimeter response can be represented as an image of the calorimeter cluster, we first applied state-of-the-art computer vision model architectures. The first preliminary experiment is the application of the fairly fast and simple ResNet model~\cite{he2016deep} which has become a cornerstone of modern computer vision due to its remarkable performance in a variety of tasks. We used neural architecture similar to the \mbox{ResNet-18} model, which is customized for the regression problem and whose convolutional layer parameters are matched to the dimensions of the input data. The resulting model has about 11.1 million of trainable parameters. The ReLU activation function is selected and the batch size, learning rate and regularization hyperparameters for the $\mathrm{RMSE}(E)$ and $\mathrm{RMSE}(E)/E$ loss functions and Dataset A are searched over a small grid of hyperparameters. Even with the best found values of the hyperparameters\footnote{The following ResNet-18 hyperparameters are selected using a grid search with cross-validation: batch size 32, learning rate 0.01 and weight decay 0.01}
of such a model, we found its performance to be significantly inferior to the baseline approach which determines the energy of the particles.

Instead of continuing to tune the hyperparameters of the ResNet model, we decided to conduct the second preliminary experiment, in which we choose the minimum possible network architecture that solves the problem given the peculiarities of the data, and try to comprehensively study the properties of models based on such neural architecture.
We chose a class of models containing up to 4 hidden layers, both fully connected and convolutional.
It turned out that models containing only fully connected layers are not able to approach the performance of the baseline models.
We manually found quite a few models consisting of one or two convolutional layers followed by one or two fully connected layers whose performance is comparable to the baseline models.
However, it turned out that most of these models are not robust to changes in the initialization of the weights.
This fact determined the focus of this paper.

Since the number of parameters in such models is much smaller than in ResNet-18 (we considered models with a maximum of 24\,000 trainable parameters), and their training speed is correspondingly high, it is possible to search through a large number of hyperparameters to find the desired model, including the choice of the activation function, regularization, and optimizer.
However, to reduce the computation, we made some preliminary experiments to reduce the choice of possible activation function, regularization, and optimizer.
Thus, for each of the selected neural architectures, we first search for the activation function under a fixed optimizer selection (Adam), this is described in Section~\ref{sec-activation_functions}.
Then for the selected best activation function, for each of the selected neural architectures, the optimizer is searched.
This subject is addressed in Section~\ref{sec-optimizers}.

\subsection{Activation Functions}\label{sec-activation_functions}
\begin{figure}[h]
\includegraphics[width=0.99\linewidth]{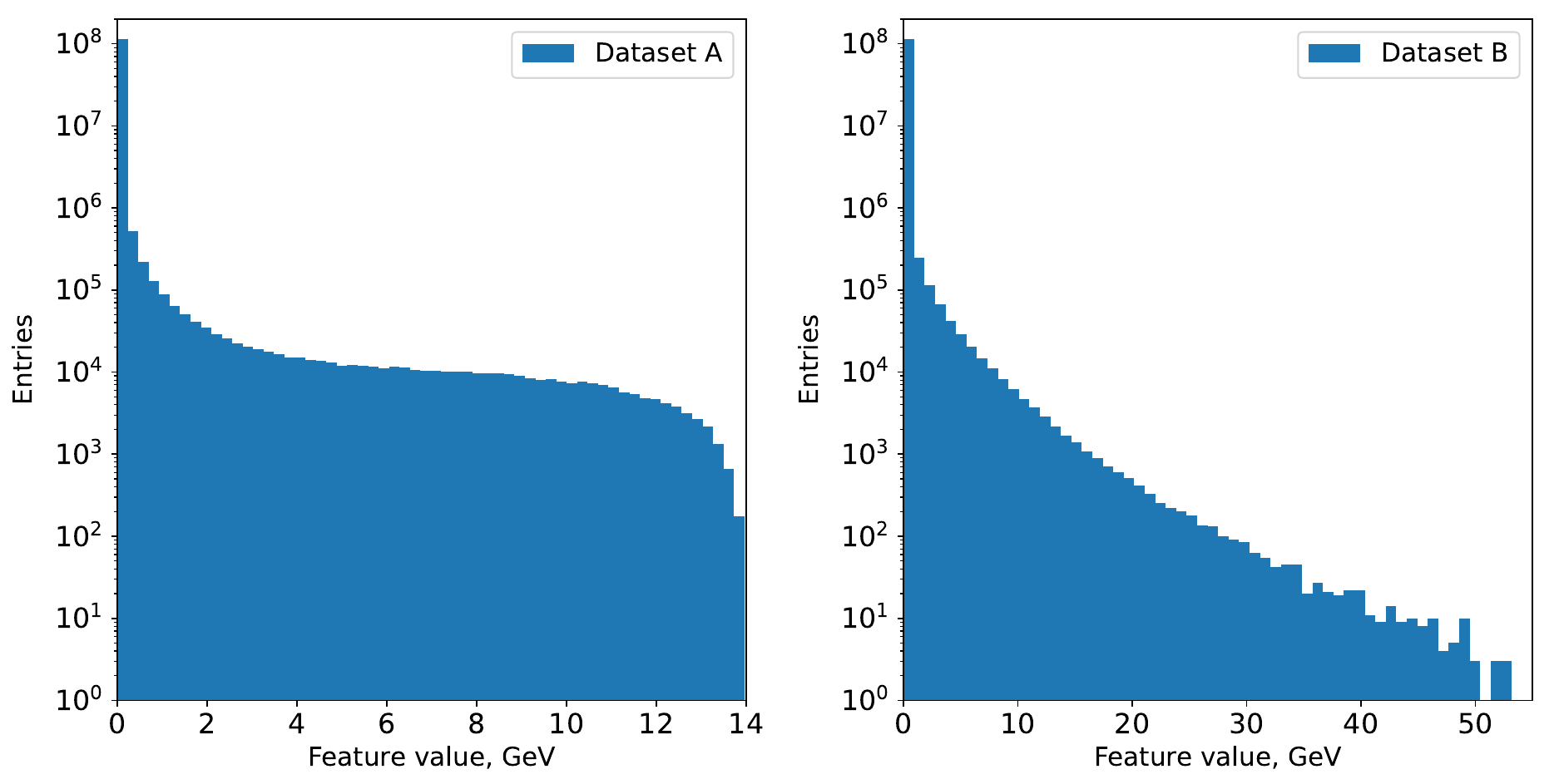}
    \caption{Histograms of the feature values in Dataset A (left) and in Dataset B (right) discussed in Section~\ref{sec-1}.}
    \label{fig_feature_values}
\end{figure}

Since the data used in the present research have very different representations in terms of feature values (as shown in Fig.~\ref{fig_feature_values}), we think it is important to discuss the well-known facts about the different activation functions used in deep learning.

An activation function is a transformation that adds non-linearity to the output of a neuron and allows the neural network to model some complex mappings between input and output data.

\begin{figure}[h] \centering \includegraphics[width=0.99\linewidth]{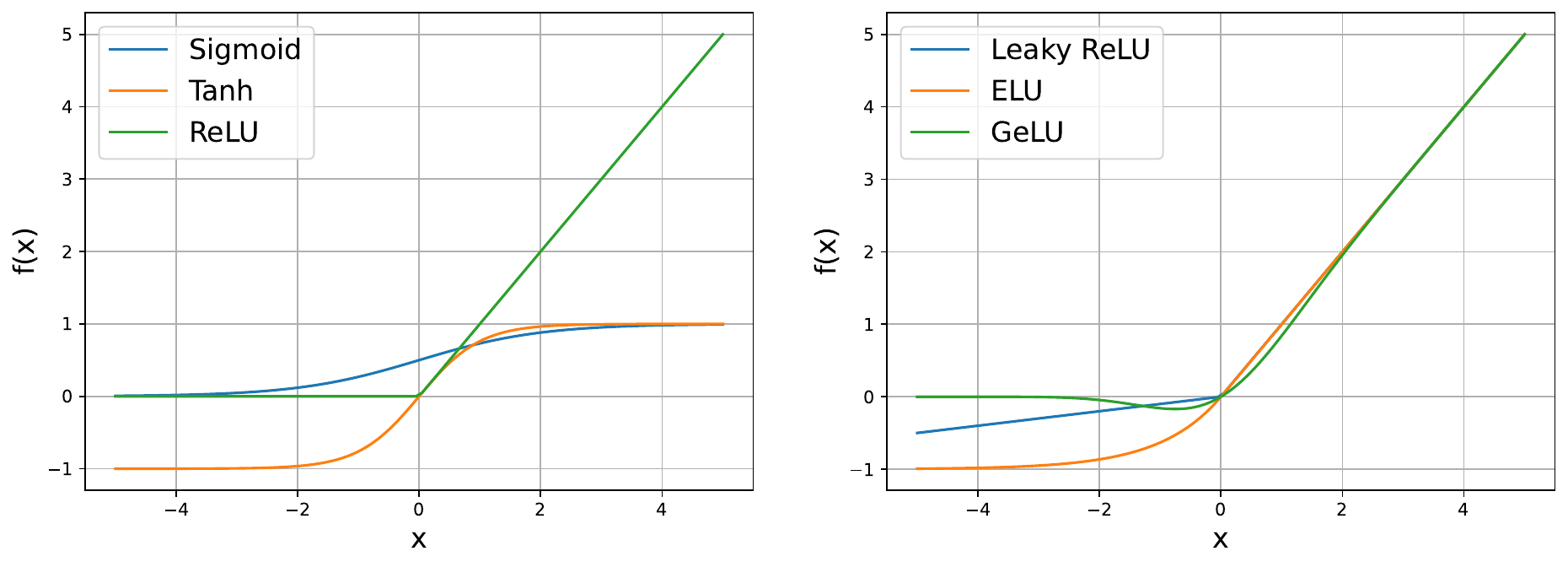} \caption{Graphs of the activation functions discussed in Section~\ref{sec-activation_functions}: Sigmoid, Tanh, ReLU (left panel), and Leaky ReLU, ELU, GELU (right panel).} \label{fig_activation_functions}
\end{figure}

It is known that the type of activation function can affect the accuracy of the model.
Thus, there are studies that show the dependence of different types of activation functions on the accuracy and training time of models~\cite{emanuel2024effect, hayou2019impact, ramachandran2017searching}.
In our study, we considered the following activation functions that are widely used: Sigmoid, Tanh, ReLU, Leaky ReLU, PReLU, ELU, and GELU. The graphs of these functions are shown in Fig.~\ref{fig_activation_functions}.

The choice of the activation function is closely related to the nature of the input data.
Thus, the interpretation of the input data within the model must not violate physical laws, for example, the energy in the cells of the calorimeter cannot take negative values, and the position of the particle in the calorimeter should follow the principles of locality.

The Sigmoid activation function converts values coming into the input to a range from 0 to 1. If the input values are large positive numbers, they will approach 1 after conversion, while negative numbers will be close to zero.
This kind of activation function may have some problems -- the vanishing of the gradient when passing through ‘saturated’ neurons.

The hyperbolic tangent (Tanh) is similar to the Sigmoid, but it converts the data to a range of $-1$ to $1$ and is zero-centered, which eliminates the vanishing gradient of the Sigmoid.

The Rectified Linear Unit (ReLU) computes the function $f(x) = \max(0, x)$, which solves the problem of zeroing the gradient for positive numbers but not for negative input data. In addition, the advantage of ReLU over Sigmoid is the speed of computation.
Leaky ReLU is a modification of ReLU that has the formula $f(x) = \max(0.01 x, x)$, which solves the problem of gradient vanishing with negative input data and still has a low computation cost. A Parametric Rectified Linear Unit (PReLU) introduced in~\cite{he2015delving} is a generalization of ReLU-flavor activation functions. PReLU allows us to learn the slope with negative input values. In this study, we applied an independent learnable parameter for PReLU for each neuron of each layer.

The Exponential Linear Unit (ELU) works in the same way as ReLU when the input is positive, and when it is negative they use the exponent, $e^x - 1$, which allows them to approximate the average unit activations to zero, similar to batch normalization, but with less computational cost~\cite{clevert2015fast}.

The Gaussian Error Linear Unit (GELU) activation function uses the formula $f(x) = x\Phi(x)$, where $\Phi(x)$ is the standard Gaussian cumulative distribution function.
As shown in ~\cite{hendrycks2016gaussian}, this activation function provides performance improvements in the  computer vision, natural language processing, and speech processing tasks.

The following experiment is performed to select the most appropriate activation function to reconstruct the particle energy.
The neural network selected for this experiment consists of two convolutional layers and two fully connected layers.
There are a total of 19\,315 trainable parameters in such a network.
Each model uses a consistent activation function chosen from the set ELU, GELU, ReLU, Leaky ReLU, PReLU, Sigmoid, or Tanh. The following hyperparameter grid is used to all these models:
  \begin{itemize}
    \item Learning rate: 0.0001, 0.001, 0.01, 0.1
    \item Batch size: 32, 64, 128, 256
  \end{itemize}

Each model is trained using the Adam optimizer~\cite{kingma2014adam} and the $\mathrm{RMSE}(E)/E$ loss function at each point of this grid for 10 times with random weight initialization and using a randomly drawn training sample. The initial sample of 64\,000 examples is taken from Dataset~A and is randomly divided into training and test sets of 32\,000 examples each\footnote{All results in this paper are based on the test samples.}. Each of these models is trained until an early stopping criterion is reached.~\footnote{The details of early stopping are described in Section~\ref{early_stopping}.}

\begin{figure}[] \centering \includegraphics[width=0.49\linewidth]{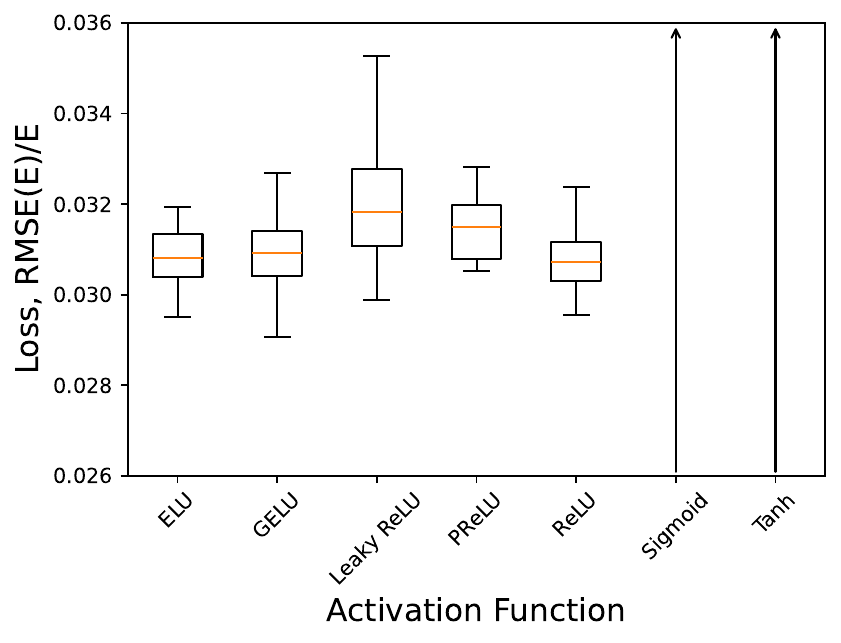} \caption{Boxplots of the losses for models consisting of 2~convolutional and 2 fully connected layers with 19\,315 trainable parameters for each activation function considered. The energy reconstruction problem is solved and Dataset A is used. 10 instances of the corresponding model are used for each box plot. Models with Sigmoid and Tanh activation functions do not fit within the range of losses shown, as indicated by the upward arrows.} \label{fig_activation_functions_results}
\end{figure}

The distribution of the losses of all instances of each model on the test sample is shown in Fig.~\ref{fig_activation_functions_results}.
For ease of comparison, they are visualized using boxplots\footnote{The variability of the losses in all boxplots in this paper is due to the use of randomly drawn training samples and different weight initializations when training the model instances. The caption for each figure with boxplot(s) indicates how many instances of the model are used for each boxplot.}.

According to the findings of the experiment, the ReLU activation function is selected, as it demonstrated best results in terms of both average loss and variability.
Further experiments showed that models using ReLU are not robust enough to determine the position of the particle, and then we used PReLU for this task.

\subsection{Optimizers}\label{sec-optimizers}

Training a neural network is an optimization problem, as the loss function is minimized in this process.
There are various analytical and numerical methods of optimization, but in ML, numerical methods are most often used.
Among the numerical methods there are direct methods which give the solution in a finite number of steps and iterative methods in which the solution is obtained as a limit when the number of iterations is increased to infinity; with a finite number of iterations approximate solutions are obtained.
All results in this paper use iterative gradient-based optimization methods, including methods that use higher-order derivatives.

There are many different optimization algorithms, each with its own advantages and disadvantages.
Due to the high computational cost, in practice only a few available optimization methods are compared and the most optimal one is chosen for a particular problem. In our work, we first considered the following methods.

Some of the most common optimization methods are stochastic gradient descent (SGD)~\cite{amari1993backpropagation} and an algorithm for first-order gradient-based optimization of stochastic objective functions based on adaptive estimates of lower-order moments (Adam)~\cite{kingma2014adam}. Paper ~\cite{hardt2016train} also discusses the robustness of stochastic gradient descent.

AdamW is a modification of the Adam optimizer that allows the recovery of the original weight decay regularization formulation by decreasing the weight from the optimization steps performed with respect to the loss function~\cite{loshchilov2017decoupled}.

The basic concept of the popular AdaGrad algorithm, which stands for "adaptive gradient", is to adjust the learning rate of each parameter over time by using the accumulated sum of all the derivatives for that parameter~\cite{duchi2011adaptive}. AdaGrad has been demonstrated to be particularly effective for sparse features where the learning rate must decrease more slowly for infrequently occurring terms.

Root mean squared propagation (RMSprop)~\cite{hinton2012neural} is an adaptive learning rate optimization algorithm that is an extension of the SGD and AdaGrad algorithms and is designed to significantly reduce the computational effort in training neural networks by exponentially decaying the learning rate every time the quadratic gradient is less than a certain threshold.

Adadelta is a method that dynamically adapts over time using only first-order information and has minimal computational cost, less than the standard SGD. The method does not require manual tuning of the learning rate and is found to be robust to noisy gradient information~\cite{zeiler2012adadelta}.

NAdam (Nesterov Momentum) is a modification of the Adam optimization algorithm in which the momentum component is implemented as a Nesterov accelerated gradient (NAG)~\cite{nesterov1983method}, which is shown to improve the rate of convergence and the quality of trained models~\cite{dozat2016incorporating}.
For batch gradient descent, Nesterov momentum can improve the rate of convergence, although for SGD it can be less effective.

The best optimizer is searched for two model architectures (2 Conv. + 2 FC and 2 Conv. + 2 FC + sum of energy deposits; both using the ReLU or PReLU activation functions) in the following grid of hyperparameters:
  \begin{itemize}
    \item Learning rate: 0.0001, 0.001, 0.01, 0.1
    \item Batch size: 8, 16, 32, 64, 128, 256
    \item L2 regularization / Weight decay: 0.0001, 0.001, 0.01, 0.1
  \end{itemize}
  
A total of 10 instances of the model are trained at each point in this hyperparameter grid.
The model instances have random initialization of weights and are trained on samples of 32\,000 examples randomly selected from a sample of 250\,000 examples.

\begin{figure}[h]
    \centering
    \includegraphics[width=0.49\linewidth]{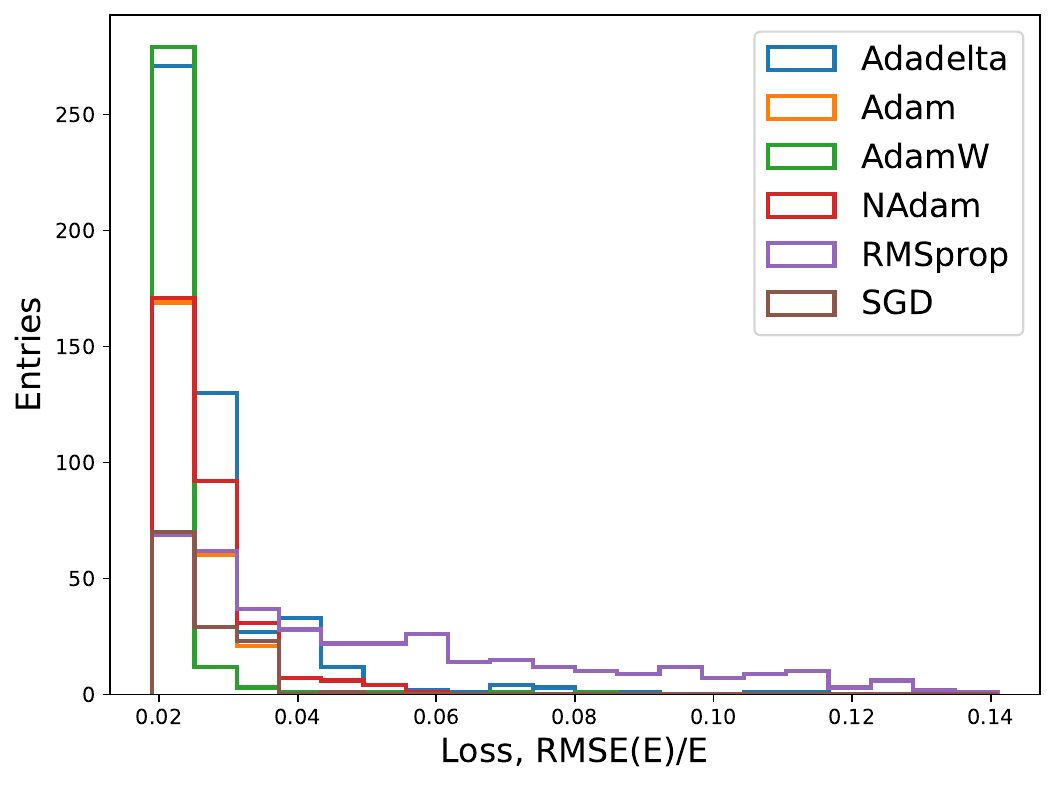}
    \caption{Histogram of the losses of all models designed for the energy reconstruction problem and discussed in Section~\ref{sec-optimizers}. Each model is trained on a sample of 32\,000 examples randomly drawn from Dataset A.}
    \label{Energy_optimizers}
\end{figure}

The results of optimizer selection are shown in Fig.~\ref{Energy_optimizers}. In further experiments, we used NAdam and AdamW, which performed best.

\subsection{Regularization}
Robust training of neural networks is closely related to exploding or vanishing gradients.
One way to deal with this is to use regularization techniques.
For our tasks, the use of batch normalization significantly speeds up learning, but does not eliminate the problem of training instability.
Preliminary experiments showed that for the particular problem and the neural architecture used, dropout regularization either does not change the robustness of model learning or makes it unpredictable.
In all experiments with the exception of those utilizing the AdamW optimizer, L2 regularization is used.
In the case of the AdamW, the authentic weight decay is used.
Another popular type of regularization, early stopping, is included in the next subsection because it is of particular importance in our study.

\subsection{Early Stopping}\label{early_stopping}
As the robustness of the model in our approach is determined by the losses of a set of model instances, we consider the losses on the test sample after reaching a given number of iterations or after reaching an early stopping criterion.
The traditional learning approach of choosing the minimum loss achieved in one of the iterations of the optimization algorithm is not considered.
This is due to the fact that it results in the model being in a local minimum of the loss function, which is not particularly deep.
It is likely that other instances of the model, starting with different initializations of the weights, will not reach such a minimum.
As a result, the set of instance losses in the aforementioned conventional training approach exhibits less predictable variability.

The following early stopping strategy is used.
All instances of the model are trained to at least epoch 100 unless otherwise stated.
Training continued until the maximum loss in the previous 30 epochs exceeded the minimum loss in the previous 30 epochs by 10\%.
This strategy enables the identification of models exhibiting explosive loss behavior as a function of training epochs.
By varying the number of mandatory training epochs, the size of the window in which the maximum and minimum losses are compared, and the value of their difference in this window, it is possible to control the tolerance of selection to the presence and magnitude of peaks in the dependence of losses as a function of training epochs.
Preliminary experiments showed that the use of different learning rate scheduling (factor and cyclical schedulers, and a warm-up) did not show the expected results in stabilizing loss behavior and therefore is not used in this study.

\subsection{Inductive Bias}\label{IB}
Following the suggestion of the authors of the paper ~\cite{goyal2022inductive} to investigate the influence of high-level variables on the learning of latent attribute representations, we have identified two inductive biases that play a causal role and can be easily introduced in the dataset we used due to its nature.
For both problems –– the reconstruction of the particle energy and the reconstruction of the particle position –– we compare the robustness and performance of models using only the input feature tensor and models that also incorporate input features calculations, as in the baseline models.
For energy reconstruction, the sum of energy deposits in all cells is passed to the network after the first fully connected layer.
To reconstruct the coordinate, the barycenter of all energy deposits is transmitted after the first fully connected layer.
Importantly, in our research these inductive biases are used in their pure form in baseline models, so comparing the training of models with inductive bias, without it, and baseline models measures the extent to which high-level variables play a causal role and how this affects model robustness.

\subsection{Implementation Details}\label{Implementation}
After the pre-selection of the activation function and loss function, the model architecture is optimized.
The following steps are taken to obtain a stable solution outside the learning cycle:
\begin{itemize}
  \item The loss function $\mathrm{RMSE}(E)/E$, $\mathrm{RMSE}(x)$ is selected;
  \item The entire available dataset is used, for which the division into training and test subsamples is fixed to 0.5/0.5;
  \item Model architecture:
  \begin{itemize}
    \item Size and number of convolution filters;
    \item Number of fully connected layers;
    \item Number of neurons in fully connected layers;
    \item Activation function;
    \item Optimizer;
    \item The use of additional input information (see Section~\ref{IB} for the details);
  \end{itemize}
  \item Model hyperparameters:
  \begin{itemize}
    \item Learning rate;
    \item Batch size;
    \item L2 regularization / Weight decay.
  \end{itemize}
\end{itemize}

The training cycle begins with the selection of the training sample and random seed selection to initialize the model weights.
The training sample is bootstrapped from a sample of 250\,000 examples (half of the entire dataset).
The weights of the neural network models are initialized using He initialization~\cite{he2015delving} in all experiments in this paper.
The model is trained until the early stopping condition is reached, see Section~\ref{early_stopping} for details.
The training of all neural networks described in this paper is done in PyTorch~\cite{paszke2019pytorch}.
The NVidia Tesla V100 GPUs provided by HPC facilities~\cite{kostenetskiy2021hpc} at HSE University are used for the core computations in PyTorch.

\subsection{Model Selection Algorithm}\label{Algorithm}
Instead of an exhaustive search of possible model architectures and hyperparameters and then analyzing whether the resulting model is robust or not, we propose to use the following robust model selection algorithm.
It generalizes the assumption that few measurements are needed to determine the robustness of a non-robust model, while many measurements are required to determine a robust model.

\begin{algorithm}[h]
\caption{Model Selection Algorithm}

\begin{algorithmic}[1]

\State Step s = 0
\State M = Initial set of models
\State Define selection criterion
\State Define the number k of instances of each model
\While{number of models in M $\geq 1$}
    \For{each model in M}
        \State Train model instance with new initialization
        \State Update model robustness value
    \EndFor
    \State s = s + 1
    \If{s $>$ k}
        \State Update selection criterion
        \For{each model in M}
            \If{model robustness is worse than\\
            selection criterion}
                \State Remove model from M
            \EndIf
        \EndFor
    \EndIf
\EndWhile
\end{algorithmic}
\label{Algorithm_1}
\end{algorithm}

\begin{figure}[h]
    \centering
    \includegraphics[width=0.49\linewidth]{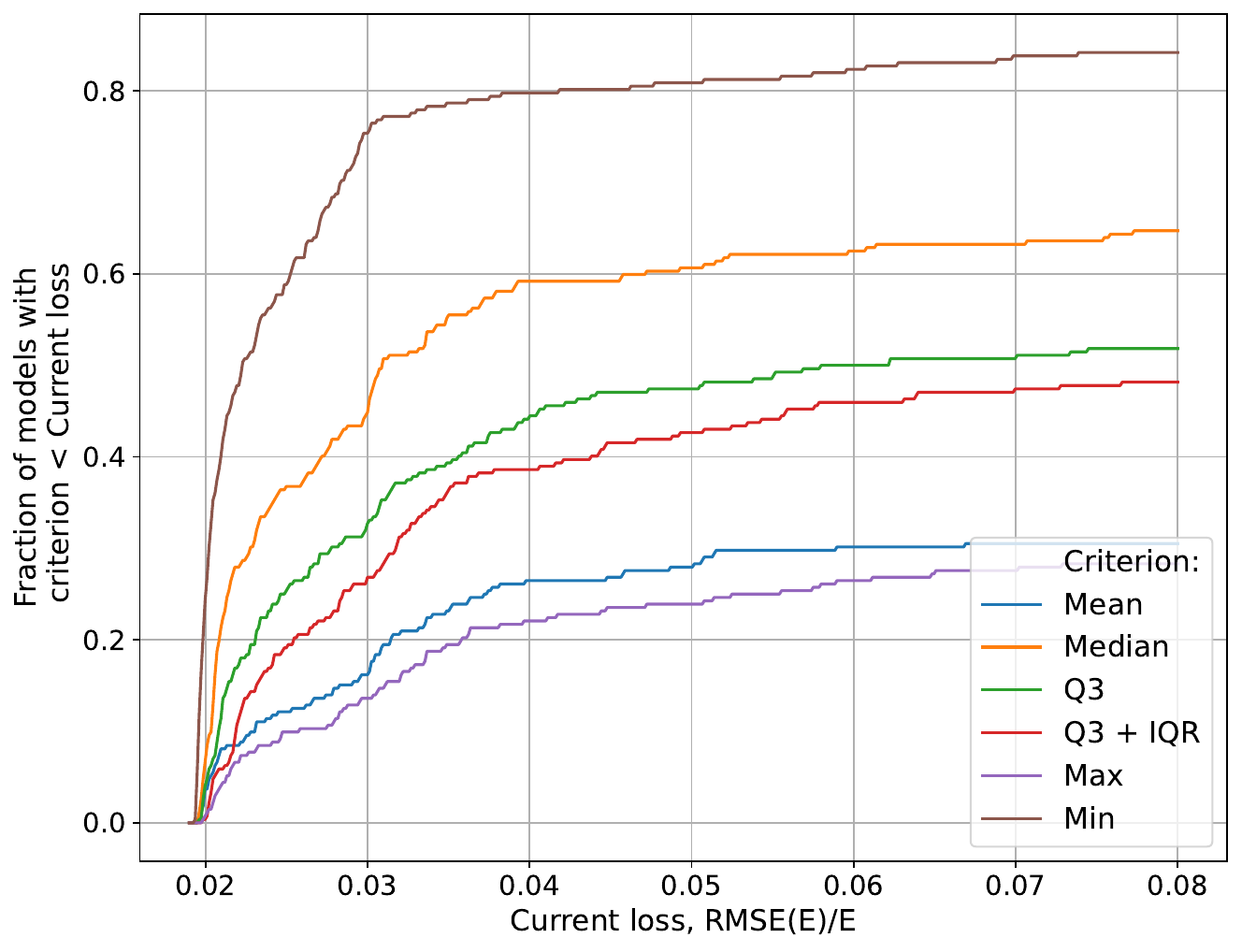}
    \caption{Line graphs of the fraction of models with losses less than the current loss for six model selection criteria. Each model aims to solve the energy reconstruction problem and is trained on a sample of size 32\,000 examples randomly drawn from Dataset A.}
    \label{Model_selection_criterion}
\end{figure}

To measure the model robustness, we propose to use one of the statistical criteria applied to a set of loss of model instances trained  independently on different input data and weight initializations.
Such statistical criteria can be: mean, median, minimum, maximum, etc. value of the loss estimated for a set of instances of the model.
The choice of model selection criterion for energy reconstruction and Dataset A is illustrated in Fig.~\ref{Model_selection_criterion}.
It can be seen that the maximum loss criterion gives the most stringent selection.
The versatility of the presented model selection algorithm lies in the fact that, depending on the selection criterion used, it is possible to select either the best performing model (in which case it will be possible to measure its robustness in the future) or a model with predetermined robustness.
Thus, using the maximum loss criterion tends to select the best performing models rather than the most robust models.
The maximum loss criterion selects the model with the best performance among all models presented at the input of the model selection algorithm.
However, if an instance of such a model produces a poor loss on average less often than the number of iterations of the model selection algorithm, then such a model may not meet the possible robustness requirements.

The search for a robust model for the energy reconstruction problem is carried out among the models that use only the input feature tensor, and among the models that also exploit the sum of the energy deposits transmitted after the first fully connected layer, as described in Section~\ref{IB}.
All models considered in the search have two convolutional layers each and from two to four fully connected layers.
Convolution filters with sizes of $2\times2$, $3\times3$, or $5\times5$ are used.
For each combination of possible model architectures, the following grid of hyperparameters is used:
  \begin{itemize}
    \item Learning rate: 0.0001, 0.001, 0.01, 0.1
    \item Batch size: 16, 32, 64, 128, 256
    \item L2 regularization / Weight decay: 0.001, 0.01, 0.1
  \end{itemize}

The combination of the above-noted architecture options and hyperparameters yields 6\,912 models.
This set of models is fed into a model selection algorithm that uses mean loss as the selection criterion.
The model instances have random initialization of weights and are trained on samples of 32\,000 examples randomly selected from 250\,000 examples of Dataset A.
To save computation, after the first training, all models whose losses exceeded the baseline loss + 20\% are excluded.
In each subsequent step of the model selection algorithm, half of the models with the highest mean loss calculated over all previous training of model instances are discarded.
It took 12 iterations of the model selection algorithm to select a robust model.
It turned out to be a model using the sum of the energy deposits transmitted after the first fully connected layer.
Also, from previous iterations of the model selection algorithm, the best model using only the input feature tensor is selected.
Subsequently, 50 more instances of each of these two models are trained to determine their robustness with higher accuracy.
Thus, a total of 41\,567 models and their instances are trained.
If the model selection algorithm is replaced by an exhaustive search for the most robust model from the set of 6\,912 models with the same robustness guarantee, then 6\,912 $\times$ 50 = 345\,600 models would have to be trained.

To find the best model for the position reconstruction problem, the same procedure is used, except that instead of the sum of energies, each model additionally receives two features corresponding to the position of the barycenter of the calorimetric cluster.
These two features are transmitted after the first fully connected layer, as described in Section~\ref{IB}. 

Although we consider two different problems, energy reconstruction and position reconstruction, and that these problems use different loss functions (as described in Section~\ref{LossFunctions}), applying model selection algorithm to the set of network architectures selects similar architectures (except for the difference in the activation functions of ReLU and PReLU, and for additional fully connected layer for the position reconstruction models) for the most robust models.

The best selected models are discussed in Section~\ref{sec-results}, their architectures are listed in Appendix~\ref{best_architectures}.
To test how the models generalize, we do model selection on the simpler Dataset A, and then train the selected models from scratch on the more complex Dataset B.
With the selected models, we conducted additional experiments to find out the factors affecting their robustness.

\subsubsection*{Dependence on training sample size}
The rationale behind this dependence is to determine the size of the training sample at which the loss value saturates and new training data is no longer beneficial.
The loss variability used to construct each boxplot is due to training each of the 50 instances of the selected model independently on different training samples and using random initialization of the model weights.
All these instances of Model 1 (Energy) and Model 2 (Energy) are independently trained on each of 45 subsamples of sizes from 100 to 32\,000 examples\footnote{The size of $i$th training sample is defined according to the formula:\\ $ 2\,000 \cdot 10^{-1.18 + i \cdot 2.38 / 44} ~~\mathrm{for}~~  i = 0, \ldots, 45$.}.
The size of the test sample matched the size of the corresponding training sample.
For each of the 45 subsamples, training and test samples are selected at random from the corresponding training and test parts of the initial data set of 500\,000 observations.

A separate comparison is also made with baseline and XGBoost\footnote{The XGBoost-based approach uses the following feature engineering: calculating the sums of energy in $3\times3$ and $5\times5$ cells surrounding the cell with maximum energy, and creating polynomial combinations of the features. The following XGBoost hyperparameters are selected using a grid search with cross-validation: \texttt{learning\_rate=0.1}, \texttt{n\_estimators=75}, \texttt{max\_depth=3}, \texttt{subsample=0.8}, \texttt{alpha=0.8}, \texttt{colsample\_bytree=0.8}. More details about this approach are described in~\cite{boldyrev2020ml}.} approaches, where all models are trained on each of samples of four different sizes: 8\,000, 16\,000, 24\,000, and 32\,000 examples.
In this experiment, 50 instances of Model 2 are trained on a randomly selected subsample of given size with different initializations of the model weights.
The source of uncertainty in all compared models is the randomness of the corresponding split between training and test samples, and in Model 2 it is also the random initialization of the weights.

\subsubsection*{Impact of model initialization}\label{Initialization}
We tried to determine how much the initialization of the model weights affects the robustness of Model 1 and Model 2.
In our problem, there are two sources of uncertainty in the loss of the chosen model: the randomness of the training and test samples, and the initialization of the model weights.
For each of the models, we conducted 3 experiments using Dataset B.
In the first experiment, 1\,000 instances of the model are trained on a fixed training sample of 32\,000 examples.
Each model instance is given a unique random seed to initialize the model weights using the He method~\cite{he2015delving}.
In the second experiment, 1\,000 model instances are trained on a randomly selected sample of 32\,000 examples (bootstrapped from a sample of 250\,000 examples).
Each model instance is given a unique random seed to initialize the sampling.
In the third experiment, 1\,000 instances of the model are trained on a random training sample of 32\,000 examples with random initialization of the model weights.

\section{Results} \label{sec-results}
\subsection{Energy Reconstruction}\label{energy_results}
Two robust models for energy reconstruction are chosen using the model selection algorithm proposed in Section~\ref{Algorithm}.
The hyperparameters of these models are summarized in Table~\ref{Table-energy}.
As can be seen in Fig.~\ref{Energy_Loss_vs_epoch}, all 50 instances of these two models trained on a sample size of 32\,000 examples converge to the 50th epoch or earlier.
It can be seen that Model 2 converges faster and requires fewer epochs to train than Model 1.
The mean training time of 50 instances of Model 1 (Model 2) is $71\pm3$ ($160\pm13$)~seconds on the NVidia Tesla V100 GPU.

\begin{table}[h]
\small
\centering
\begin{tabular}{|c|c|c|c|c|c|}
\hline
Model & \begin{tabular}[c]{@{}c@{}}Activation\\ function\end{tabular} & Optimizer & \begin{tabular}[c]{@{}c@{}}Learning\\ rate\end{tabular} & Batch size & Regularization \\ \hline
1 (Energy) & \multirow{2}{*}{ReLU} & NAdam & 0.0001 & 64 & 0.01 (L2) \\ \cline{1-1} \cline{3-6} 
2 (Energy) &  & AdamW & 0.001 & 32 & \begin{tabular}[c]{@{}c@{}}0.1\\ (weight decay)\end{tabular} \\
\hline
\end{tabular}
\caption{Hyperparameters of the models used for energy reconstruction in Datasets A and B. Both models utilize an architecture of 2 convolutional and 2 fully connected layers. Model 2 additionally uses the sum of energies transferred after the first fully connected layer. The architecture of each selected model is summarized in the Appendix~\ref{best_architectures}.} \label{Table-energy}
\end{table}

\begin{figure*}[h!]
\centering
\includegraphics[width=0.99\linewidth]{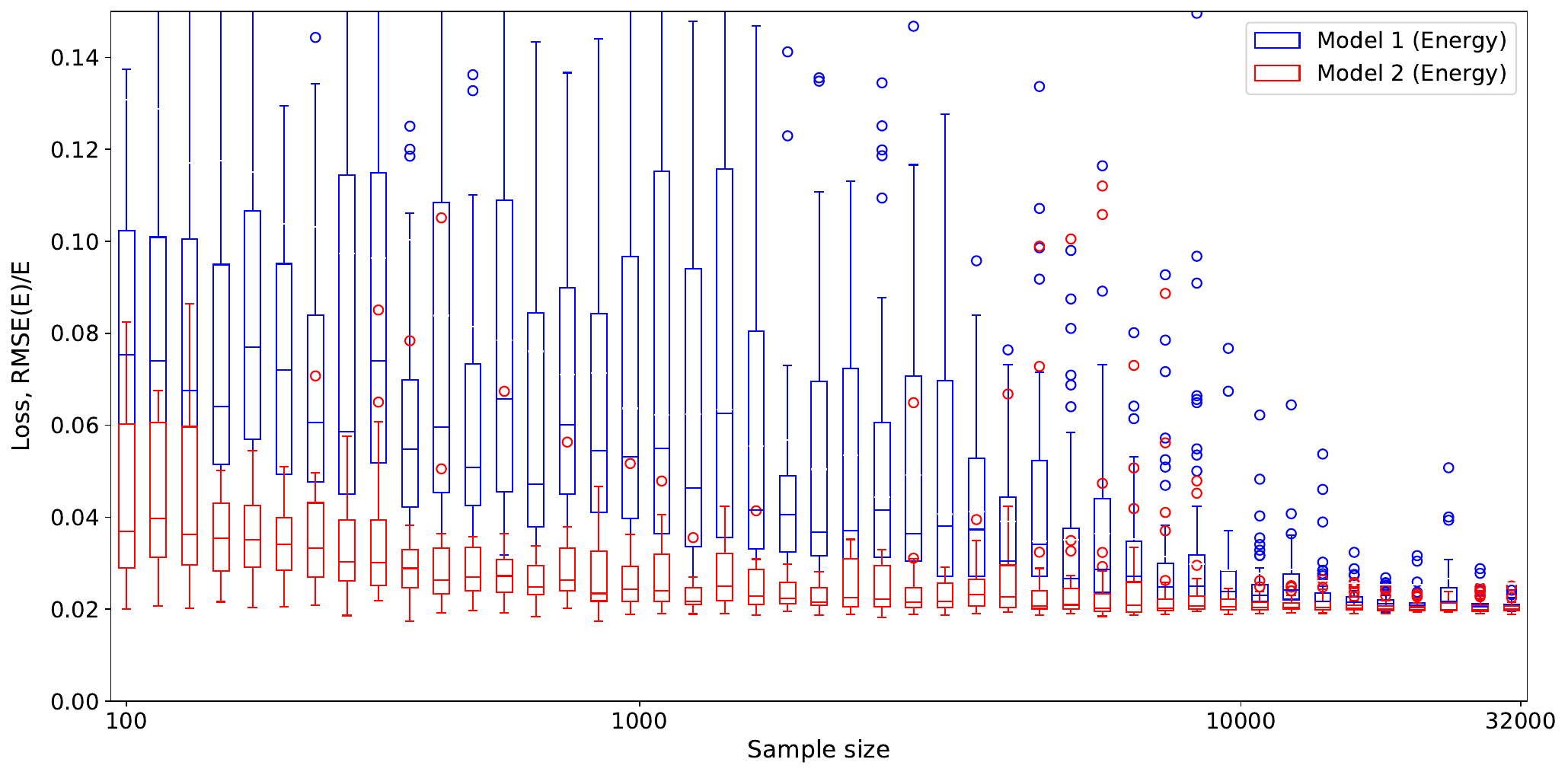}
\caption{Boxplots of the losses for the energy reconstruction problem for Model 1 and Model 2 as a function of the training sample size for Dataset A. Model 2 uses the sum of the energies in the cells transferred after the first convolution layer. 50 instances of the corresponding model are used for each box plot.} \label{Energy_A_132-32k}
\end{figure*} 

\begin{figure*}[h!]
\centering
\includegraphics[width=0.99\linewidth]{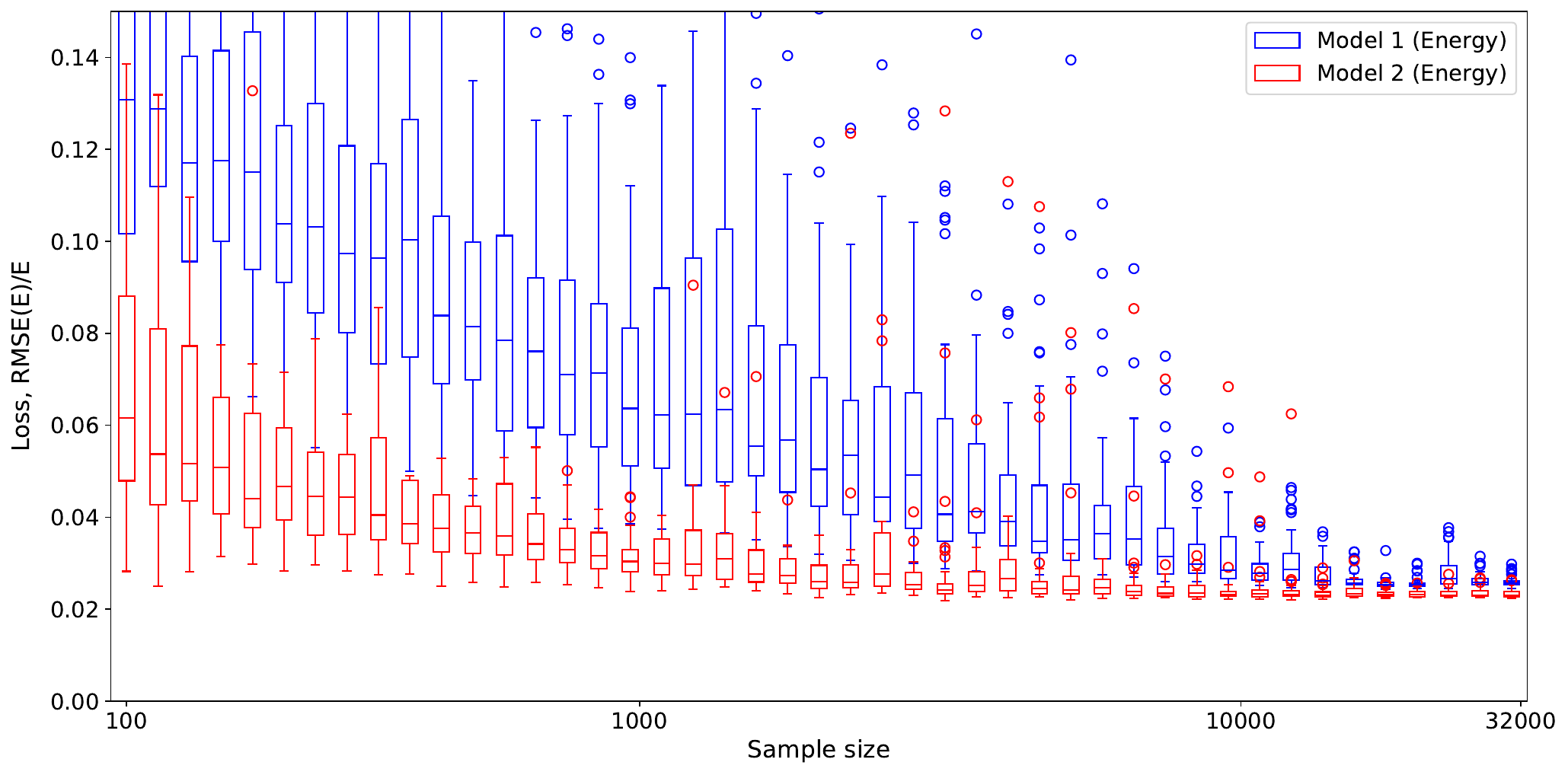}
\caption{Boxplots of the losses for the energy reconstruction problem for Model 1 and Model 2 as a function of the training sample size for Dataset B. Model 2 uses the sum of the energies in the cells transferred after the first convolution layer. 50 instances of the corresponding model are used for each box plot.} \label{Energy_B_132-32k}
\end{figure*}

Figures~\ref{Energy_A_132-32k} and~\ref{Energy_B_132-32k} show boxplots of the $\mathrm{RMSE}(E)/E$ loss on test sample for different training sample sizes for the models listed in Table~\ref{Table-energy}.
It can be seen that as the sample size increases, the variability of the loss decreases on average, and the median loss decreases almost equally.
This fits the case described by the authors of the paper~\cite{schaeffer2023double}, where double descent does not occur and we see no noticeable change in loss behavior as the ratio of the number of model parameters to the training sample size crosses the interpolation threshold.
It can be seen in Fig.~\ref{Energy_A_132-32k} that when trained on extremely small training sample sizes (100 observations), for almost all instances of Model 1 shows a higher value of the loss on the test sample than Model 2.
As it can be seen in Fig.~\ref{Energy_B_132-32k}, the difference between the losses of Model 1 and Model 2 is even higher for the more complex Dataset B.
To obtain the same level of both loss and its variability, a model trained with inductive bias (Model 2), can be trained on smaller sample sizes than a model using only raw features (Model 1).
The minimum sample size for the model to be considered robust is 18\,000 (2\,000) examples for Model 1 (Model 2).

\begin{figure}[h]
    \centering
    \includegraphics[width=0.49\textwidth]{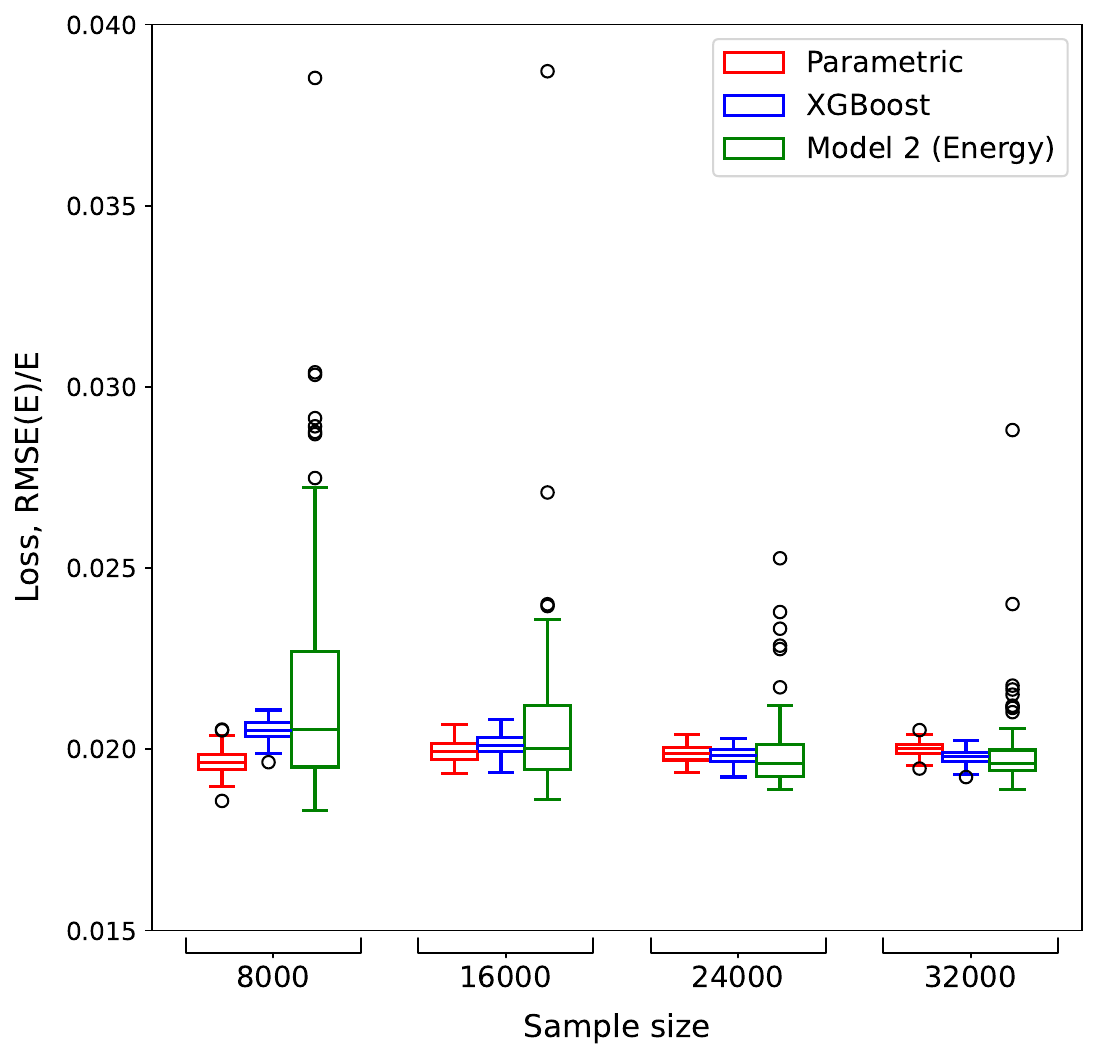}
    \includegraphics[width=0.49\textwidth]{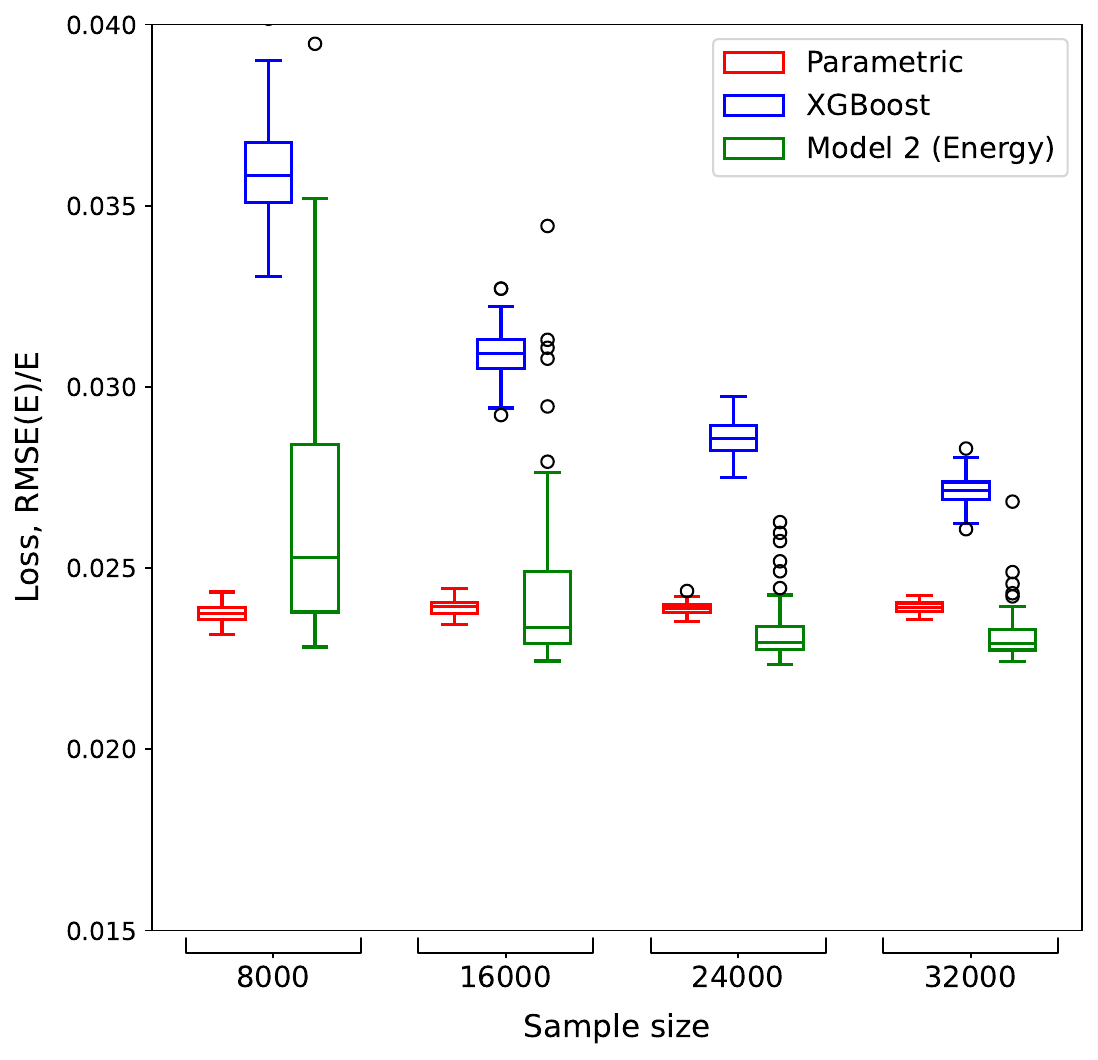}
    \caption{Boxplots of the losses for the energy reconstruction problem for the parametric approach (red), XGBoost (blue), and Model 2 (green), trained on samples of size 8\,000, 16\,000, 24\,000 and 32\,000 examples, respectively, randomly taken from Dataset A (top panel) or from Dataset B (bottom panel). 100 instances of the corresponding model are used for each boxplot.}
    \label{Energy_Dataset_4_bins}
\end{figure}

Fig.~\ref{Energy_Dataset_4_bins} shows the comparison of the boxplots of $\mathrm{RMSE}(E)/E$ loss for Model 2 with the parametric model and with the XGBoost model for four training sample sizes of Dataset A (top panel) and Dataset B (bottom panel).
It can be seen that for both data sets, for all training sample sizes, Model 2 is less robust than the parametric model.
This suggests that the set of model architectures used as input to the model selection algorithm is insufficient to find a deep learning model that is more robust than the baseline parametric model.
It can also be seen that the average loss of Model~2 is the best compared to the parametric model and XGBoost for both datasets, training on sample sizes of 24\,000 and 32\,000 examples.

\begin{figure}[h]
    \centering
    \includegraphics[width=0.49\textwidth]{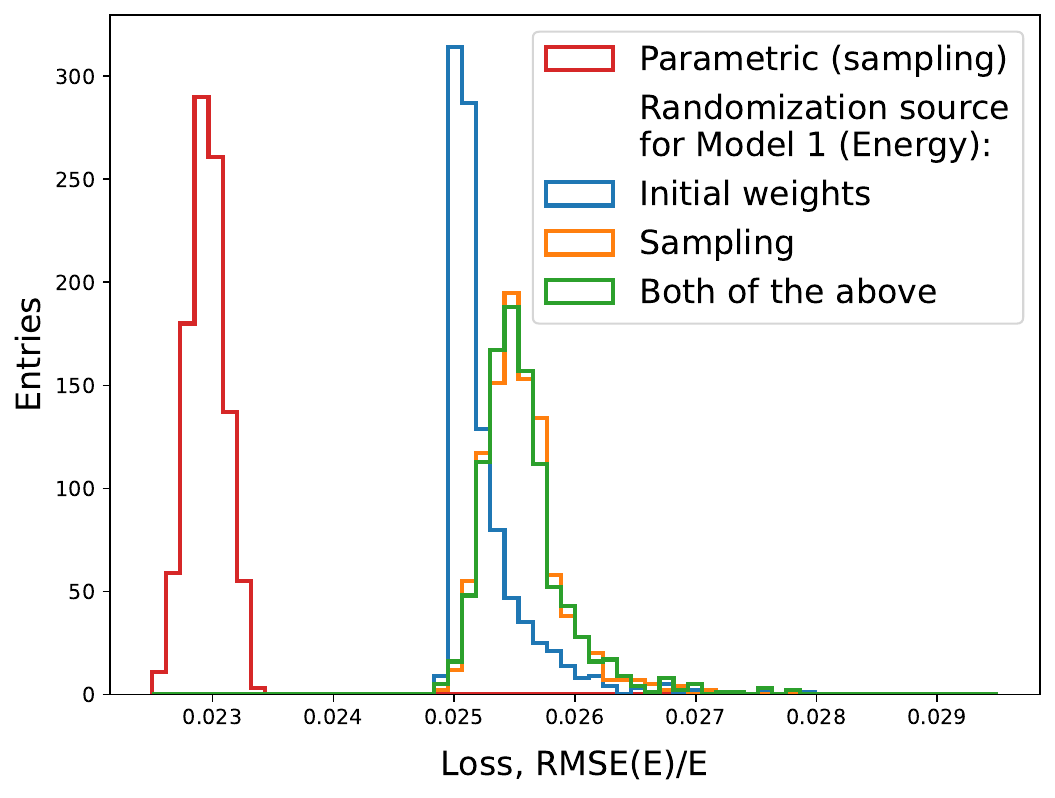}
    \includegraphics[width=0.49\textwidth]{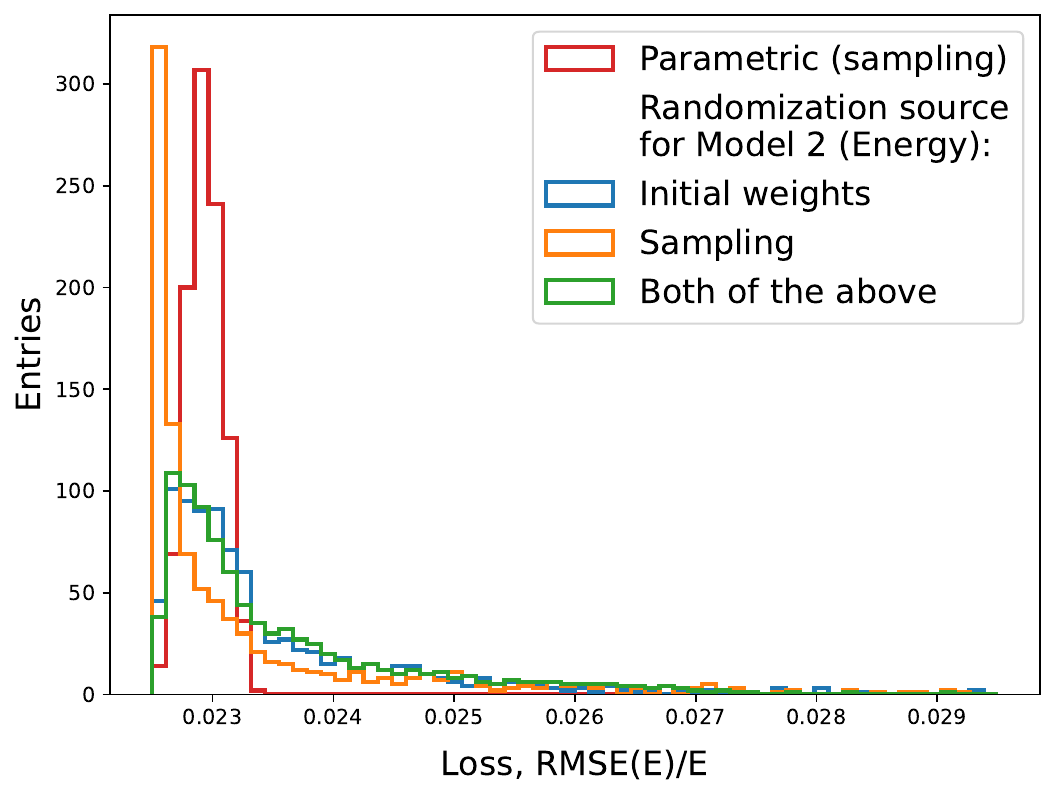}
    \caption{Histograms of the losses of different instances of Model 1 and Model 2. The top (bottom) panel shows the distribution of losses of instances of Model 1 (Model 2) depending on the choice of randomization source. Each of the 1,000 instances of the model is trained on a sample of 32\,000 examples randomly drawn from Dataset B. For comparison, both panels show the estimation of a fully parametric (baseline) approach, where the only source of uncertainty is the randomness of the partitioning into training and test samples.}
    \label{Energy_randomization_sources}
\end{figure}

The results of three experiments on the impact of weight initialization on the robustness of selected models are shown in Fig.~\ref{Energy_randomization_sources}.
The results on the top panel show that the sampling of the training set has a slightly greater effect on the robustness of Model 1 than the effect of the initialization of the model weights.
The results on the bottom panel show that sampling of the training sample affects the robustness of Model 2 slightly weaker than the effect of the initialization of the model weights.
This effect is less pronounced than for Model 1.

\subsection{Position Reconstruction}\label{position_results}
\begin{figure}[h]
    \centering
    \includegraphics[width=0.99\linewidth]{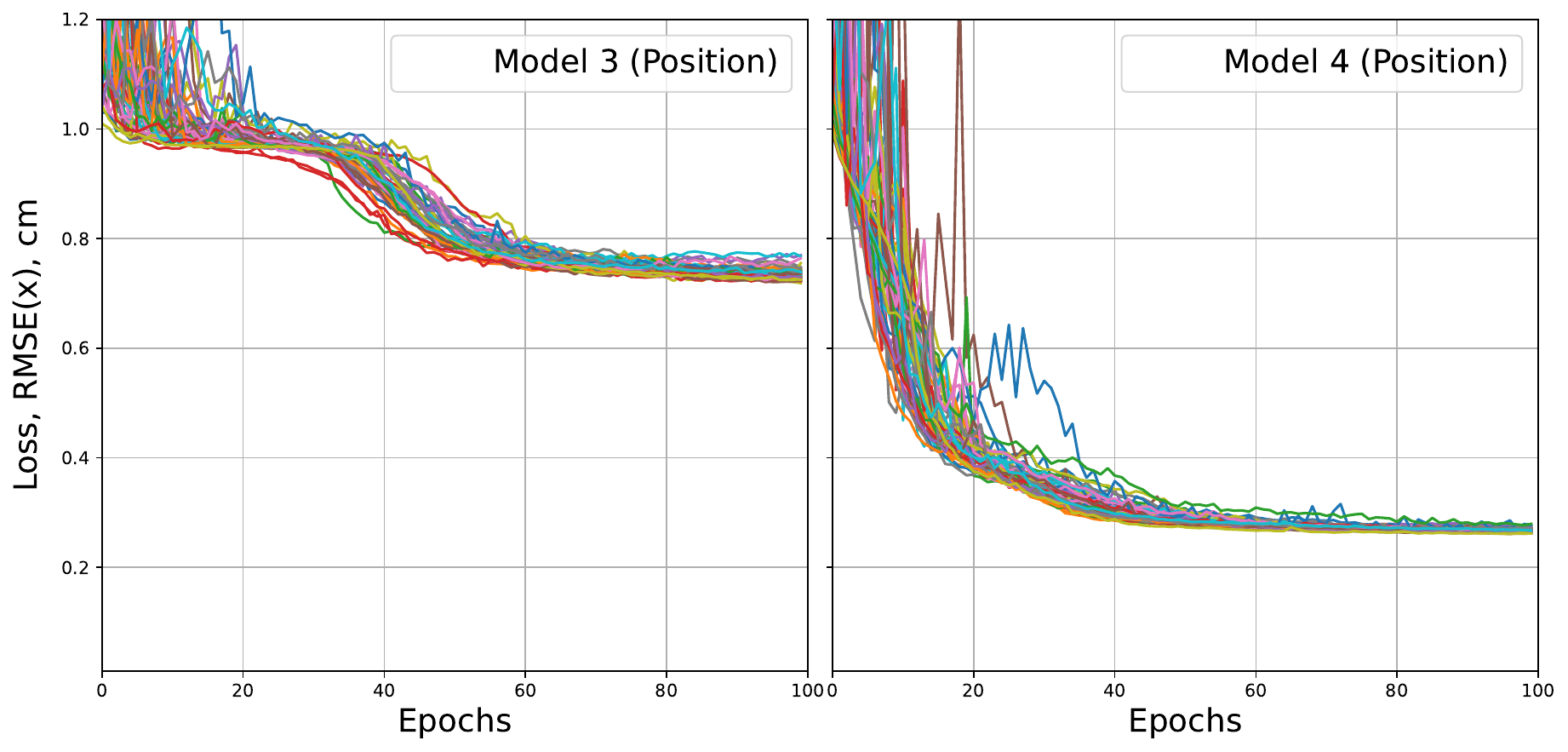}
    \caption{Evolution of the loss of the instances of Model 3 (left panel) and Model 4 (right panel) models as a function of training epochs. Different colors indicate different instances of the corresponding model. Each model aims to solve the position reconstruction problem and is trained on a sample of size 32\,000 examples randomly drawn from Dataset B.}
    \label{Position_Loss_vs_epoch}
\end{figure}

The models aimed to reconstruct the position of the particle are chosen using Algorithm 1.
The selected robust models consist of two convolutional and three fully connected layers.
Their hyperparameters are listed in Table~\ref{Table-position}.
A model that uses only raw features is called Model 3 (Position).
A model extended with the information about the barycenter position is called Model 4 (Position).
In this model, the barycenter position of energy deposits is passed after the first fully connected layer, as described in Section~\ref{IB}.
As can be seen in Fig.~\ref{Position_Loss_vs_epoch}, all 50 instances of these two models trained on a sample size of 32\,000 examples converge to the 100th epoch or earlier.
The mean training time of 50 instances of Model 3 (Model 4) is $300\pm6$ ($430\pm7$)~seconds on the NVidia Tesla V100 GPU.

\begin{table}[h]
\small
\centering
\begin{tabular}{|c|c|c|c|c|c|}
\hline
Model & \begin{tabular}[c]{@{}c@{}}Activation\\ function\end{tabular} & Optimizer & \begin{tabular}[c]{@{}c@{}}Learning\\ rate\end{tabular} & Batch size & Regularization \\ \hline
\begin{tabular}[c]{@{}c@{}}3 (Position)\\ 4 (Position)\end{tabular} & PReLU & AdamW & 0.0001 & 32 & \begin{tabular}[c]{@{}c@{}}0.1\\ (weight decay)\end{tabular} \\
\hline
\end{tabular}
\caption{Hyperparameters of the models used for position reconstruction in Datasets A and B. Both models utilize an architecture of two convolutional and three fully connected layers and have the same hyperparameters. Model 4 additionally uses the barycenter position transferred after the first fully connected layer. The architecture of each selected model is summarized in the Appendix~\ref{best_architectures}.} \label{Table-position}
\end{table}

\begin{figure*}[h]
\centering
\includegraphics[width=0.99\linewidth]{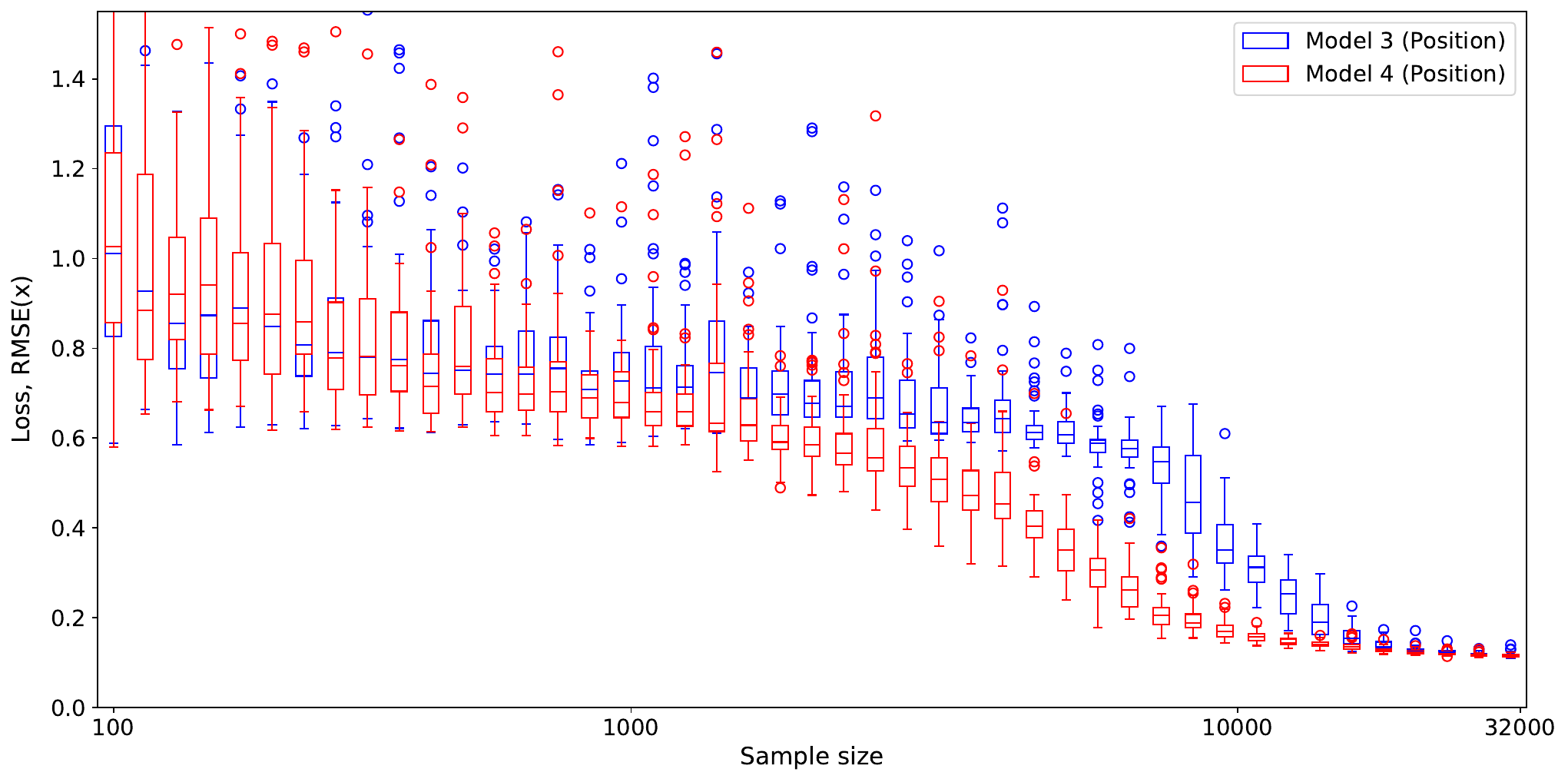}
\caption{Boxplots of the losses for the position reconstruction problem for Model 3 and Model 4 as a function of the training sample size for Dataset A. Model 4 uses the barycenter position of the energy deposits after the first fully connected layer. 50 instances of the corresponding model are used for each box plot.} \label{Position_A_132-32k}
\end{figure*}

\begin{figure*}[h]
\centering
\includegraphics[width=0.99\linewidth]{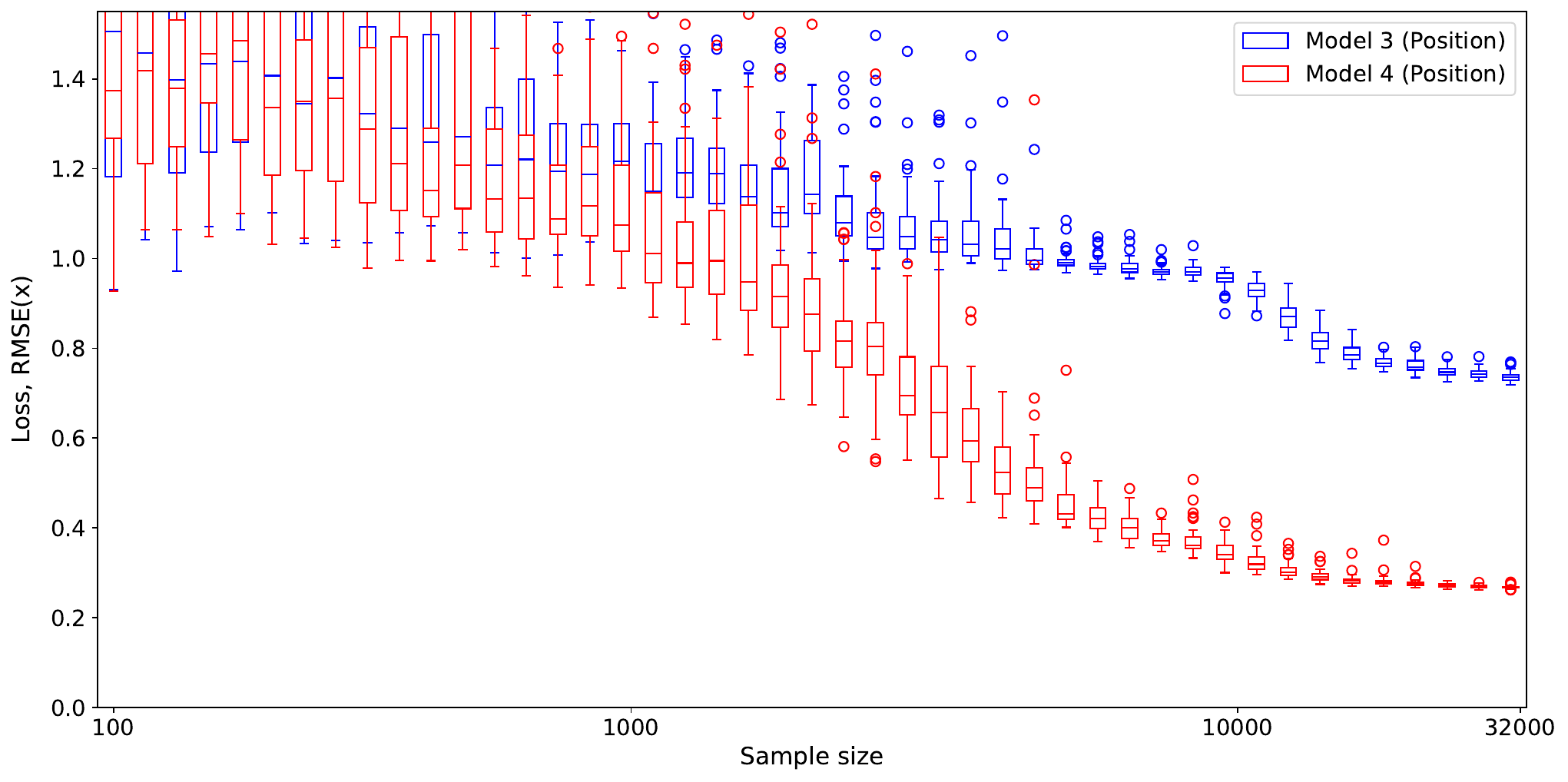}
\caption{Boxplots of the losses for the position reconstruction problem for Model 3 and Model 4 as a function of the training sample size for Dataset B. Model 4 uses the barycenter position of the energy deposits after the first fully connected layer. 50 instances of the corresponding model are used for each box plot.} \label{Position_B_132-32k}
\end{figure*}

\begin{figure}[h]
    \centering
    \includegraphics[width=0.49\textwidth]{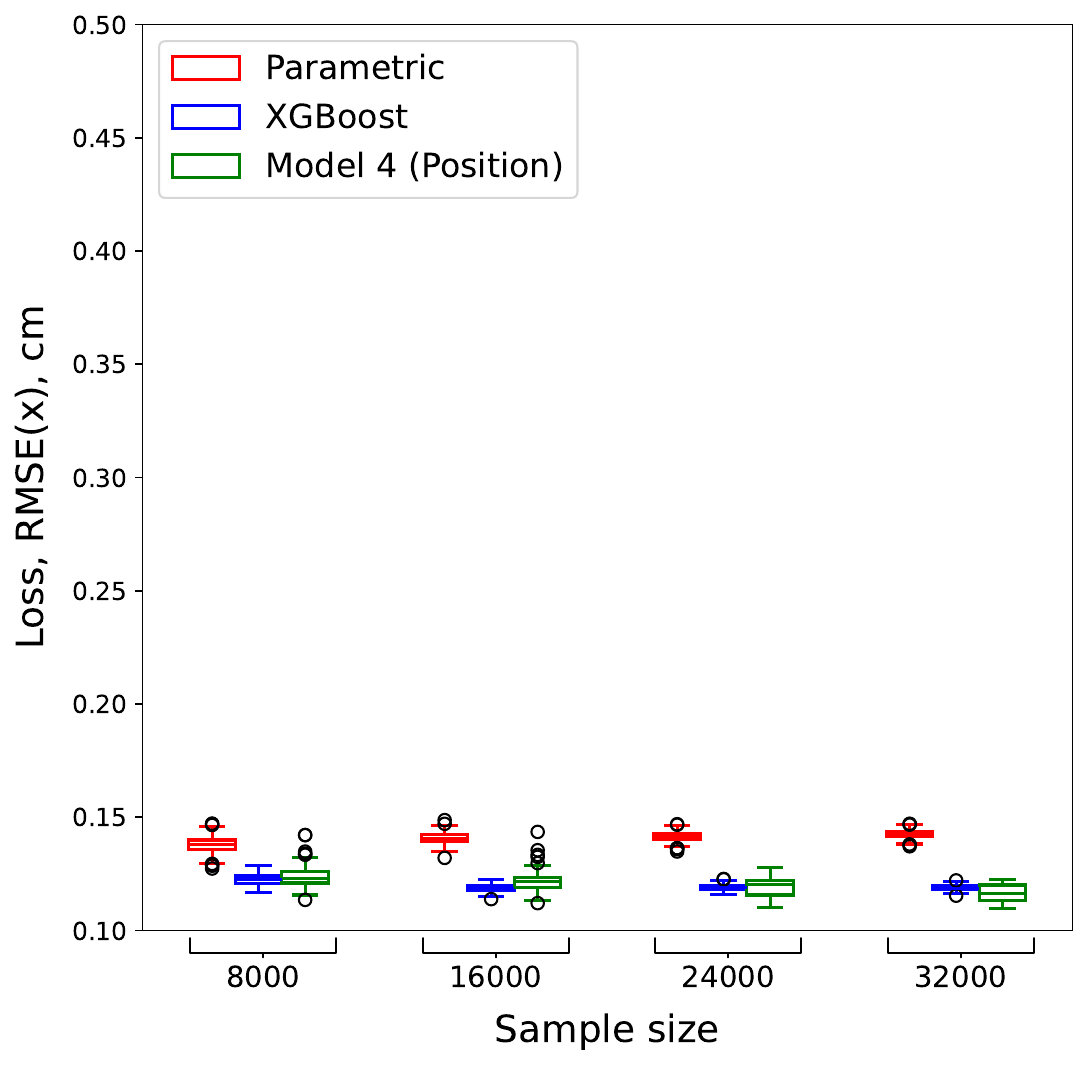}
    \includegraphics[width=0.49\textwidth]{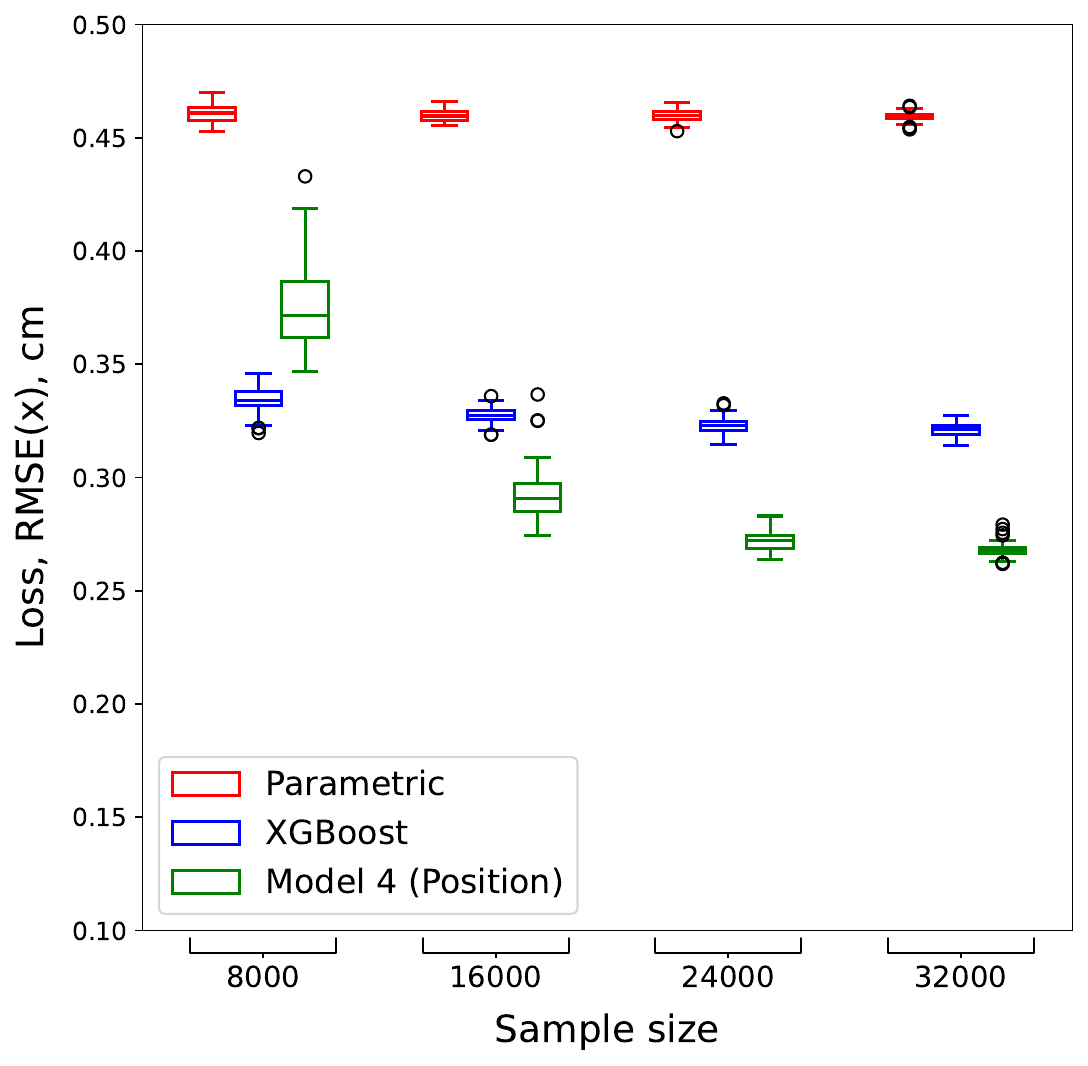}
    \caption{Boxplots of the losses for the position reconstruction problem for the parametric approach (red), XGBoost (blue), and Model 4 (green), trained on samples of size 8\,000, 16\,000, 24\,000 and 32\,000 examples, respectively, randomly taken from Dataset A (top panel) or from Dataset B (bottom panel). 100 instances of the corresponding model are used for each boxplot.}
    \label{Position_Dataset_4_bins}
\end{figure}

The Figures~\ref{Position_A_132-32k},~\ref{Position_B_132-32k},~\ref{Position_Dataset_4_bins} are obtained in the same way as the corresponding Figures~\ref{Energy_A_132-32k},~\ref{Energy_B_132-32k},~\ref{Energy_Dataset_4_bins} of the subsection~\ref{energy_results}.
As it can be seen in Figures~\ref{Position_A_132-32k} and~\ref{Position_B_132-32k}, the variability of the losses decreases on average and its average values stabilizes with further increase in sample size for both Model 3 and Model 4.
Both the losses and their variability decrease steadily until the training sample size reaches approximately 8\,000 examples. Beyond this point, unlike models addressing the energy reconstruction problem, the loss decreases by more than threefold for sample sizes ranging from 9\,000 to 15\,000 examples. During this range, the variability in loss is substantially higher compared to both smaller and larger sample sizes. This effect is visually accentuated when using a logarithmic scale on the horizontal axis of Figure~\ref{Position_A_132-32k}, although it is also observable using a linear scale. After reaching a sample size of around 16\,000 examples, both the loss and its variability stabilize. At this sample size, Model 3 can be considered robust. Notably, Model 3 narrows the performance gap with Model 4 (denoted by red-colored boxplots in the same figure) when trained with sample sizes between 9\,000 and 15\,000 examples. This suggests that this is the minimum data volume required for Model 3 to effectively assimilate information typically provided to Model 4 directly from the input tensor. Therefore, a model equipped with inductive bias, like Model 4, can be adequately trained on smaller sample sizes compared to a model relying solely on raw features, such as Model 3.

For the more complex Dataset B, Model 3 also shows a gradual decrease in loss while training on samples in the range from 9\,000 to 15\,000 examples, as shown in Figure~\ref{Position_B_132-32k}.
In this case, the drop in loss does not exceed a factor of $1.3$.
Comparing Model 3 and Model 4, it can be seen that the capacity of Model 3 is insufficient to assimilate the information that Model 4 receives directly from the input tensor.
Starting from a sample size of about 16\,000 examples, the loss and its variability decrease slowly.

In contrast to the energy reconstruction problem, where two models converged to close loss when trained on a large sample in both datasets, here such behavior is observed only for Dataset A on samples of 17\,000 or more examples.
This value can be considered as the minimum size for the training sample size for a robust model.
For Dataset B, the mean loss of Model 3 cannot come close to the mean loss of Model 4.

For Model 4, the loss and its variability decrease almost uniformly with increasing training sample size for both Dataset A and Dataset B, stabilizing at a sample size of about 16\,000 (this value can be considered as the minimum training sample size for a robust model).
The average loss for a sample size of 32\,000 examples is 2.3 times higher than for Dataset B.

The top panel of Figure~\ref{Position_Dataset_4_bins} shows that for a training sample size of 32\,000 examples, the loss of Model 4 is less than that of the other models, and the robustness of Model 4 is worse than that of the other models.
The bottom panel of Figure~\ref{Position_Dataset_4_bins} shows that for sample sizes greater than 16\,000 examples, the loss of Model 4 is less than that of the other models, and for a sample size of 32\,000 examples, it also shows comparable robustness to the other models.

\section{Conclusions}\label{sec_conclusions}
Finding a reliable model is a desirable property for many machine learning applications.
In fault-tolerant scenarios, finding robust deep learning models is particularly important.
Our study proposes a new approach for determining the robustness of machine learning models and an algorithm for selecting robust models from a set.
The two robust models for two specific problems found using this method have the best convergence and the smallest loss variability among the 2$\times$6\,912 models considered.
Finding a robust model from a set of 6\,912 models with different architectures and hyperparameters requires training 41\,567 models and their instances, instead of training 345\,600 models and their instances while considering an exhaustive search for the most robust model from the set.
For the models found, we determined the minimum training sample size at which such models can be considered robust.
We found that when we consider the robust model using only raw features, the uncertainty factor of the sampling of the training set dominates the uncertainty factor of the random initialization of the model weights.
The opposite is true for a model with inductive bias, but the effect is less pronounced.
It was also found that the robust model using high-level variables as the inductive bias requires a slightly smaller training sample, converges faster, has smaller losses, but its robustness does not change compared to the model without inductive bias.

The proposed robust model selection algorithm is a meta-algorithm and can work effectively for automated model training, enhancing existing AutoML approaches.

Future research may focus on applying the proposed approach to determine the robustness of very deep neural networks and state-of-the-art models on different datasets.
We also consider it useful to investigate possible selection criteria in the proposed model selection algorithm, including those that allow the selection of models with guaranteed robustness.

\clearpage
\section{Appendix}
\subsection{Architecture of selected models}\label{best_architectures}
The following summaries of the architectures of the best performing models are obtained using the torchinfo library~\cite{Yep_torchinfo_2020}. The model name is on the first line of the "Layer" column.

\begin{minipage}{\linewidth}
\small
\begin{lstlisting}
================================================
Layer              Output Shape      	Param #
================================================
Model 1 (Energy)  [-1]               	--
|--Conv2d         [-1, 32, 13, 13]   	288
|--ReLU           [-1, 32, 13, 13]   	--
|--MaxPool2d      [-1, 32, 6, 6]     	--
|--Conv2d         [-1, 64, 4, 4]     	18,432
|--ReLU           [-1, 64, 4, 4]     	--
|--MaxPool2d      [-1, 64, 3, 3]     	--
|--Linear         [-1, 9]           	5,193
|--ReLU           [-1, 9]           	--
|--Linear         [-1, 1]            	10
================================================
Trainable params: 23,923
================================================
\end{lstlisting}
\end{minipage}

\begin{minipage}{\linewidth}
\small
\begin{lstlisting}
================================================
Layer              Output Shape      	Param #
================================================
Model 2 (Energy)  [-1]               	--
|--Conv2d         [-1, 32, 13, 13]   	288
|--ReLU           [-1, 32, 13, 13]   	--
|--MaxPool2d      [-1, 32, 6, 6]     	--
|--Conv2d         [-1, 64, 4, 4]     	18,432
|--ReLU           [-1, 64, 4, 4]     	--
|--MaxPool2d      [-1, 64, 3, 3]     	--
|--Linear         [-1, 9]           	5,202
|--ReLU           [-1, 9]           	--
|--Linear         [-1, 1]            	10
================================================
Trainable params: 23,932
================================================
\end{lstlisting}
\end{minipage}

\begin{minipage}{\linewidth}
\small
\begin{lstlisting}
================================================
Layer              Output Shape      	Param #
================================================
Model 3 (Position)  [-1]               	--
|--Conv2d           [-1, 32, 13, 13]   	288
|--PReLU            [-1, 32, 13, 13]   	32
|--MaxPool2d        [-1, 32, 6, 6]     	--
|--Conv2d           [-1, 64, 4, 4]     	18,432
|--PReLU            [-1, 64, 4, 4]     	64
|--MaxPool2d        [-1, 64, 1, 1]     	--
|--Linear           [-1, 64]           	4,160
|--PReLU            [-1, 64]           	64
|--Linear           [-1, 9]            	585
|--PReLU            [-1, 9]            	9
|--Linear           [-1, 1]            	10
================================================
Trainable params: 23,644
================================================
\end{lstlisting}
\end{minipage}

\begin{minipage}{\linewidth}
\small
\begin{lstlisting}
================================================
Layer              Output Shape      	Param #
================================================
Model 4 (Position)  [-1]               	--
|--Conv2d           [-1, 32, 13, 13]   	288
|--PReLU            [-1, 32, 13, 13]   	32
|--MaxPool2d        [-1, 32, 6, 6]     	--
|--Conv2d           [-1, 64, 4, 4]     	18,432
|--PReLU            [-1, 64, 4, 4]     	64
|--MaxPool2d        [-1, 64, 1, 1]     	--
|--Linear           [-1, 64]           	4,160
|--PReLU            [-1, 64]           	64
|--Linear           [-1, 9]            	603
|--PReLU            [-1, 9]            	9
|--Linear           [-1, 1]            	10
================================================
Trainable params: 23,662
================================================
\end{lstlisting}
\end{minipage}

\subsection{Supplementary images}

\begin{figure}[h]
    \centering
    \includegraphics[width=0.49\linewidth]{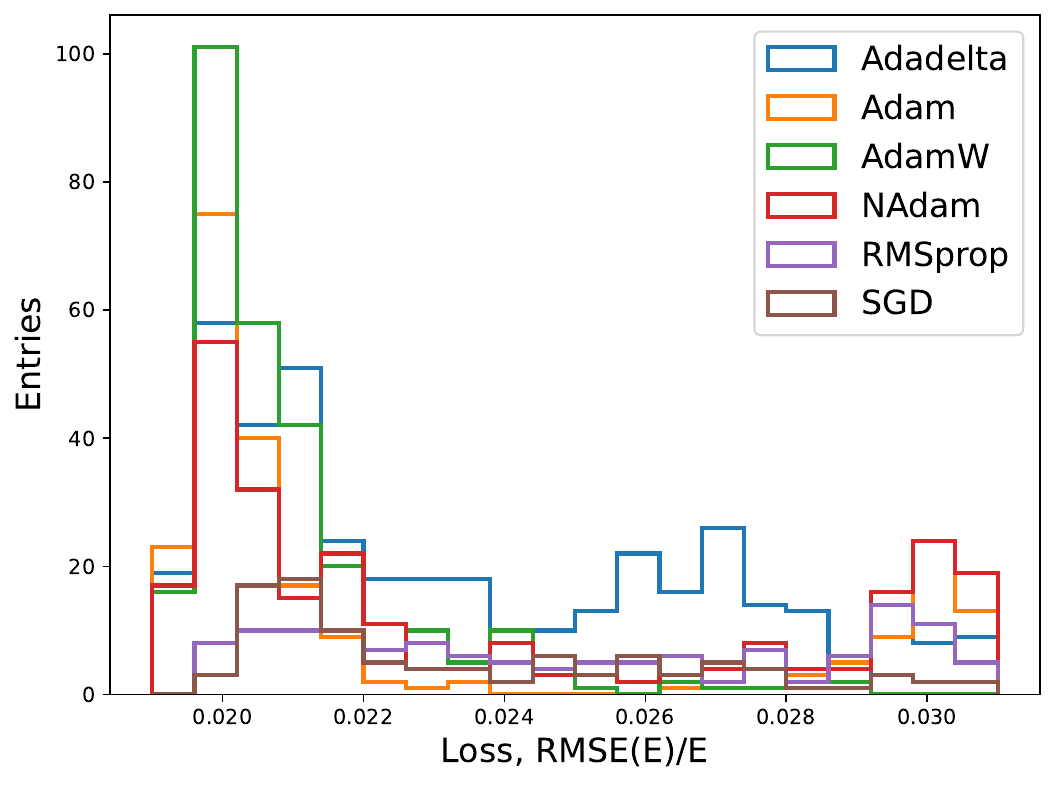}
    \caption{Histogram of the losses of all models designed for the energy reconstruction problem and discussed in Section~\ref{sec-optimizers}. Each model is trained on a sample of 32\,000 examples randomly drawn from Dataset A. This is the zoomed and rebinned version of Fig.~\ref{Energy_optimizers}.}
\end{figure}

\begin{figure}[h]
    \centering
    \includegraphics[width=0.99\linewidth]{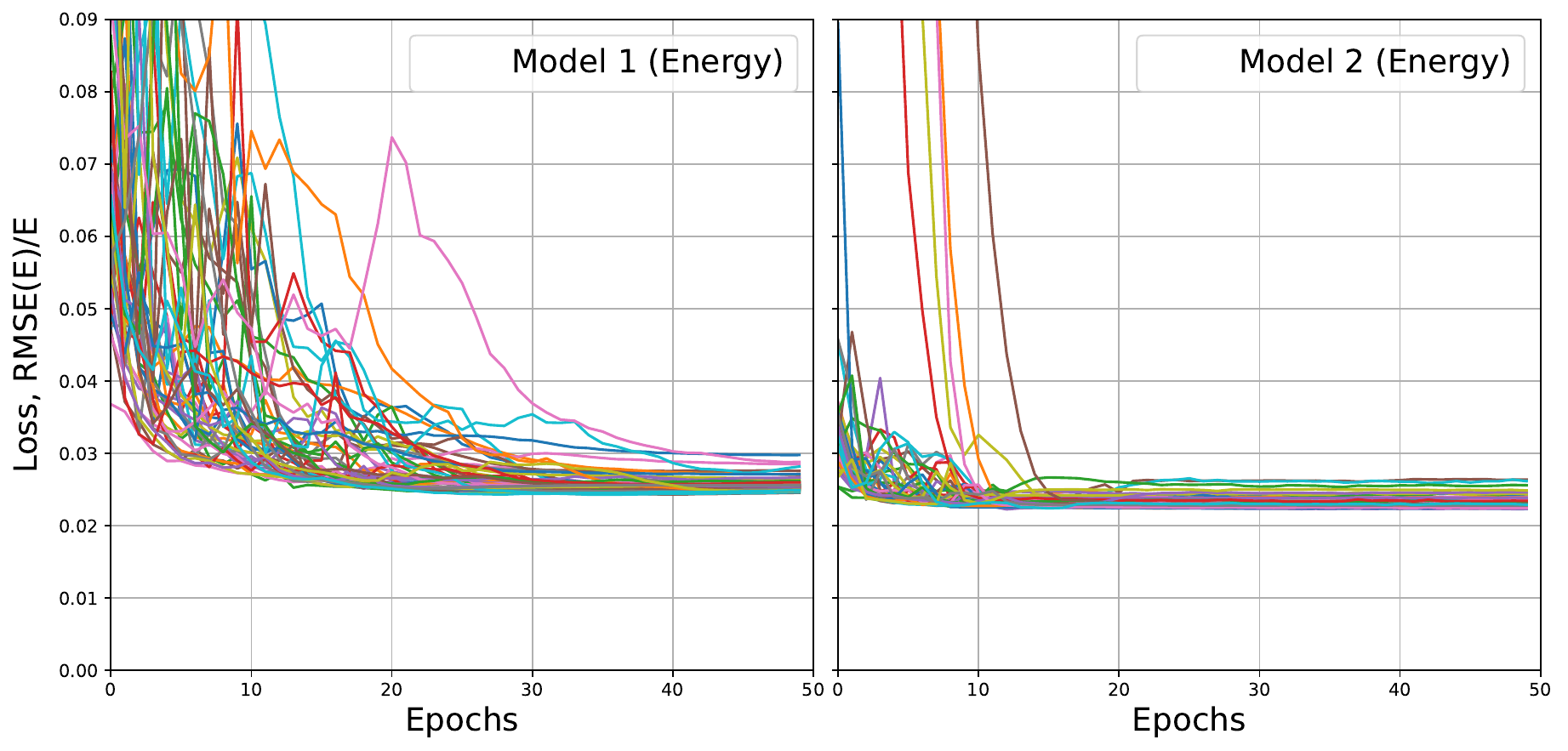}
    \caption{Evolution of the loss of the instances of Model 1 (left panel) and Model 2 (right panel) models as a function of training epochs. Different colors indicate different instances of the corresponding model. Each model aims to solve the energy reconstruction problem and is trained on a sample of size 32\,000 examples randomly drawn from Dataset B.}
    \label{Energy_Loss_vs_epoch}
\end{figure}

\clearpage
\section{Comparison with NAS}\label{NAS_comparison}
Below are the implementation details of the NAS approach applied to the search space of neural network architectures limited to up to 2 convolutional and up to 3 fully connected layers,
and to the energy reconstruction problem using Dataset B. The same loss function ($\mathrm{RMSE}(E)/E$) and training sample of the same size (32\,000 examples) are used as in the robust model selection approach described in Section~\ref{Algorithm}. The NAS from the Optuna library is used~\cite{akiba2019optuna}. This tool provides the ability to search multiple hyperparameters using the Tree-structured Parzen Estimator algorithm~\cite{watanabe2023tree} to select hyperparameter values.

The NAS uses the following set of hyperparameters:
  \begin{itemize}
    \item Number of filters in each convolution layer independently: 16, 32, 64
    \item Number of hidden fully connected layers: 1, 2, 3
    \item Number of neurons in fully connected layers: 32, 64, 128
    \item Optimizers: Adam, AdamW, RMSProp
    \item Learning rate: 0.0001, 0.001, 0.01
    \item Batch size: 16, 32, 64
    \item L2 regularization / Weight decay: 0.001, 0.01, 0.1
  \end{itemize}

In total, the space for optimization and selection of the best model is 9\,477 combinations of hyperparameters. It is worth noting that the hyperparameters of the Model 2, which is selected as the most robust model in our approach, are included in this set of hyperparameters.
The NAS algorithm is first run for 120 trials, resulting in the selection of the model referred to as NAS 1, followed by a run for the next 240 trials, resulting in the model referred to as NAS 2. The summaries of the NAS 1 and NAS 2 models are listed below, and the Table~\ref{Table-NAS} lists their hyperparameters.

\begin{table}[h]
\small
\centering
\begin{tabular}{|c|c|c|c|c|c|}
\hline
Model & \begin{tabular}[c]{@{}c@{}}Activation\\ function\end{tabular} & Optimizer & \begin{tabular}[c]{@{}c@{}}Learning\\ rate\end{tabular} & Batch size & \begin{tabular}[c]{@{}c@{}}Regularization\\ (weight decay)\end{tabular} \\ \hline
NAS 1 & \multirow{2}{*}{ReLU} & \multirow{2}{*}{AdamW} & 0.001 & \multirow{2}{*}{32} & 0.001 \\ \cline{1-1} \cline{4-4} \cline{6-6}
NAS 2 &  &  & 0.01 & & \begin{tabular}[c]{@{}c@{}}0.01\end{tabular} \\
\hline
\end{tabular}
\caption{Hyperparameters of the models NAS 1 and NAS 2.} \label{Table-NAS}
\end{table}

\begin{minipage}{\linewidth}
\small
\begin{lstlisting}
================================================
Layer              Output Shape      	Param #
================================================
NAS 1             [-1]               	--
|--Conv2d         [-1, 32, 13, 13]   	288
|--ReLU           [-1, 32, 13, 13]   	--
|--MaxPool2d      [-1, 32, 6, 6]     	--
|--Conv2d         [-1, 16, 4, 4]     	4,608
|--ReLU           [-1, 16, 4, 4]     	--
|--MaxPool2d      [-1, 16, 1, 1]     	--
|--Linear         [-1, 32]           	544
|--ReLU           [-1, 32]           	--
|--Linear         [-1, 9]           	306
|--ReLU           [-1, 9]           	--
|--Linear         [-1, 1]            	10
================================================
Trainable params: 5,756
================================================
\end{lstlisting}
\end{minipage}

\begin{minipage}{\linewidth}
\small
\begin{lstlisting}
================================================
Layer              Output Shape      	Param #
================================================
NAS 2             [-1]               	--
|--Conv2d         [-1, 32, 13, 13]   	288
|--ReLU           [-1, 32, 13, 13]   	--
|--MaxPool2d      [-1, 32, 6, 6]     	--
|--Conv2d         [-1, 16, 4, 4]     	4,608
|--ReLU           [-1, 16, 4, 4]     	--
|--MaxPool2d      [-1, 16, 1, 1]     	--
|--Linear         [-1, 128]           	2,176
|--ReLU           [-1, 128]           	--
|--Linear         [-1, 64]           	8,256
|--ReLU           [-1, 64]           	--
|--Linear         [-1, 9]           	594
|--ReLU           [-1, 9]           	--
|--Linear         [-1, 1]            	10
================================================
Trainable params: 15,932
================================================
\end{lstlisting}
\end{minipage}

\begin{figure}[h]
    \centering
    \includegraphics[width=0.99\linewidth]{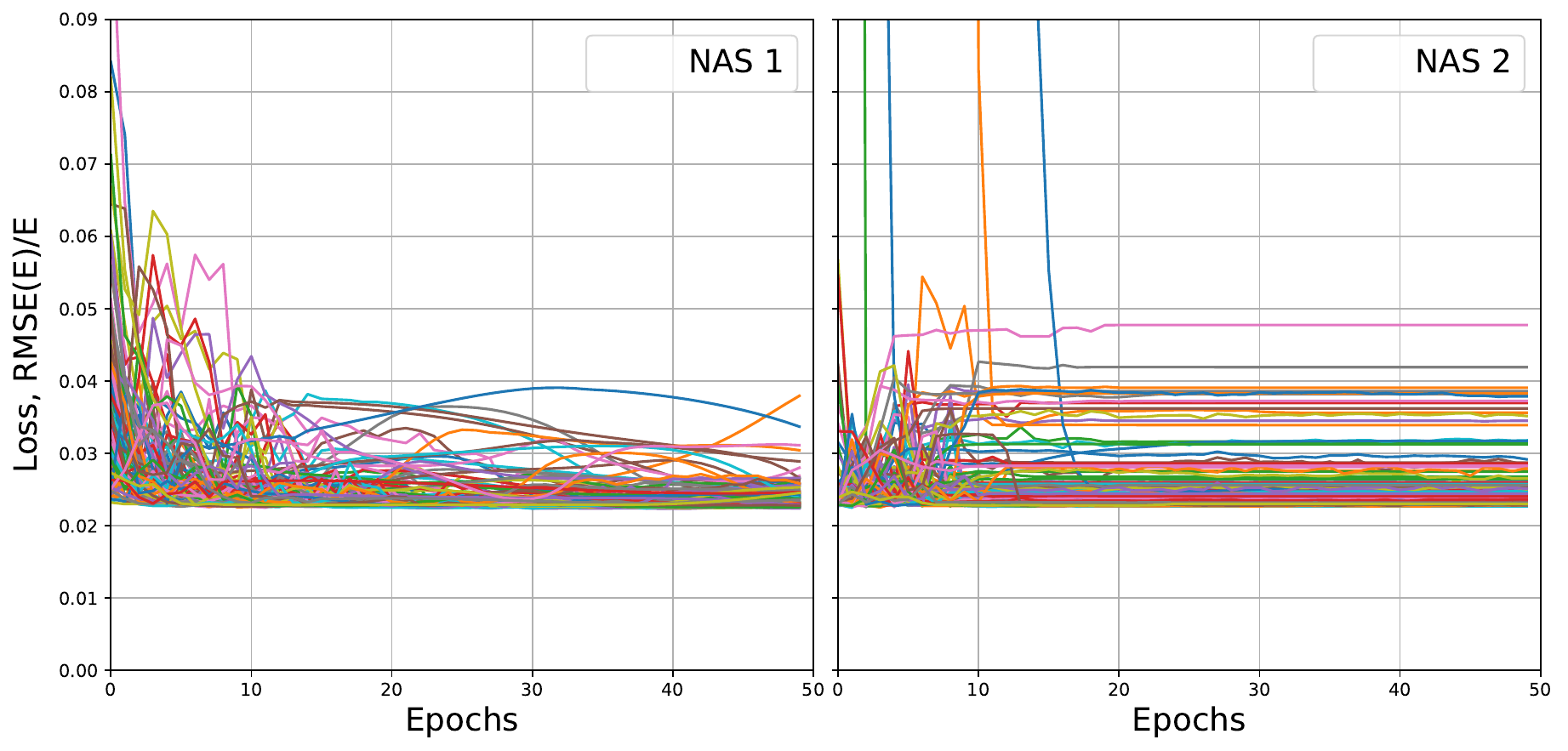}
    \caption{Evolution of the loss of the instances of NAS 1 (left panel) and NAS 2 (right panel) models as a function of training epochs. Different colors indicate different instances of the corresponding model. Each model aims to solve the energy reconstruction problem and is trained on a sample of size 32\,000 examples randomly drawn from Dataset B.}
    \label{Energy_Loss_vs_epoch_NAS}
\end{figure}

For comparison with Model 2, 50 instances of each of the NAS 1 and NAS 2 models were trained. The same methodology was used as before to obtain the results from Section~\ref{energy_results}. The Figure~\ref{Energy_Loss_vs_epoch_NAS} shows the dependence of loss on training epochs for the NAS 1 and NAS 2 models, each of whose 50 instances was independently trained on a random sample size of 32\,000 examples. This figure should be compared to the Fig.~\ref{Energy_Loss_vs_epoch} for Model 2. It can be seen that although the best losses of NAS 1 and NAS 2 instances are as good as the best losses of the Model 2 instances, the variability of their losses is greater than the variability of the Model 2 losses.

\begin{figure}[h]
    \centering
    \includegraphics[width=0.49\textwidth]{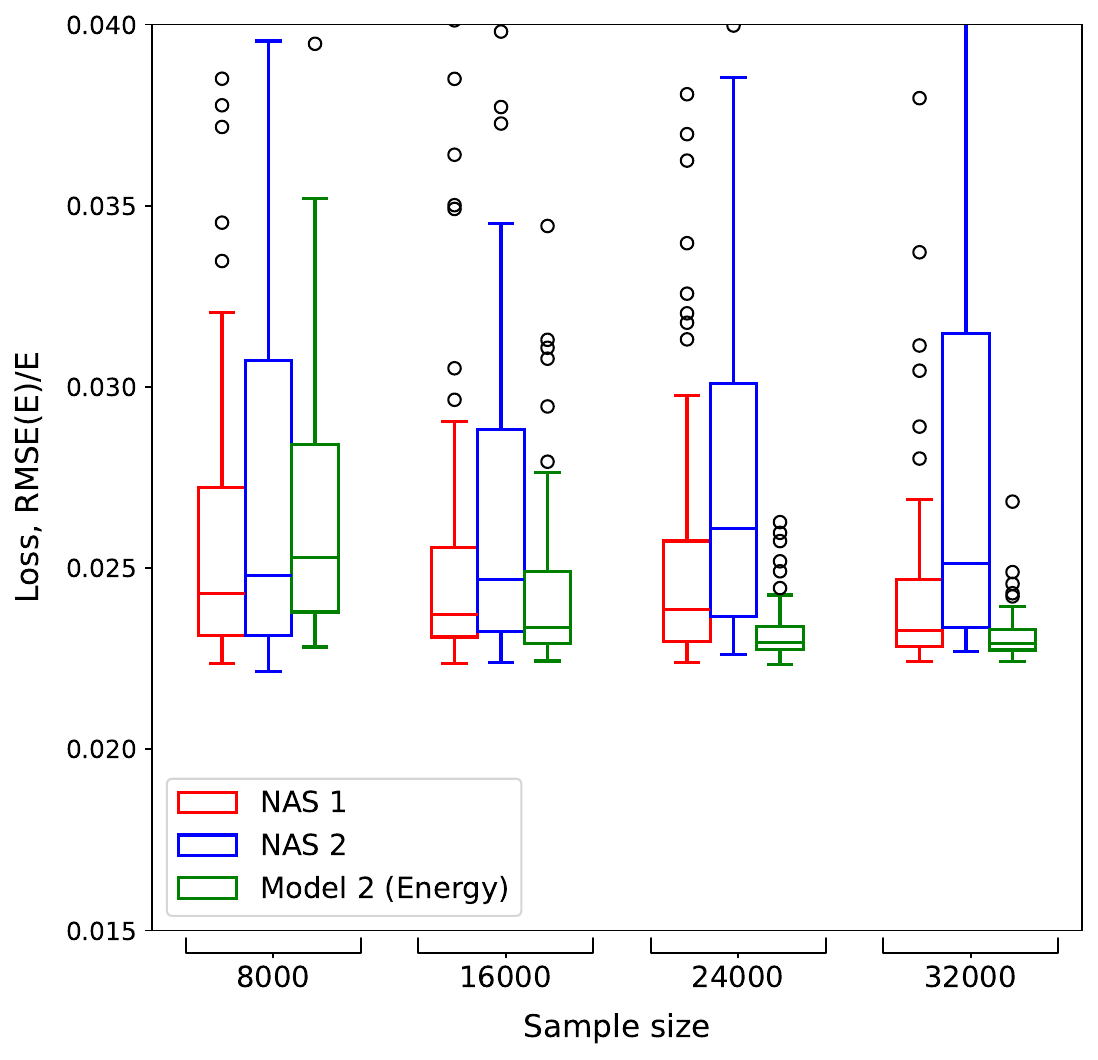}
    \caption{Boxplots of the losses for the energy reconstruction problem for NAS 1 (red), NAS 2 (blue), and Model 2 (green), trained on samples of size 8\,000, 16\,000, 24\,000 and 32\,000 examples, respectively, randomly taken from Dataset B. 100 instances of the corresponding model are used for each boxplot.}
    \label{Energy_Dataset_4_bins_NAS}
\end{figure}

Fig.~\ref{Energy_Dataset_4_bins_NAS} shows the comparison of the boxplots of $\mathrm{RMSE}(E)/E$ loss for NAS 1 and NAS 2 models with the Model 2 for Dataset B. It can be seen that Model 2, trained on samples of 16\,000 or more examples, is more robust than models NAS 1 and NAS 2.

\section*{Acknowledgment}
The publication was supported by the grant for research centers in the field of AI provided by the Analytical Center for the Government of the Russian Federation (ACRF) in accordance with the agreement on the provision of subsidies (identifier of the agreement 000000D730321P5Q0002) and the agreement with HSE University No. 70-2021-00139.

This research is supported in part through computational
resources of HPC facilities at HSE University~\cite{kostenetskiy2021hpc}.

\bibliographystyle{unsrt}  
\bibliography{ref}






\end{document}